\documentclass[preprint,sort&compress,12pt]{elsarticle}

\usepackage{algorithm}
\usepackage{algorithmicx}
\usepackage{algpseudocode}
\usepackage{array}
\usepackage{amsfonts,amsthm}
\usepackage{amsmath,amssymb}
\usepackage{multirow,booktabs}
\newtheorem{theorem}{Theorem}
\newtheorem{lemma}{Lemma}
\newproof{pf}{Proof}

\algrenewcommand\algorithmicrequire{\textbf{Input:}}
\algrenewcommand\algorithmicensure{\textbf{Output:}}

\begin{document}
\begin{frontmatter}

\title{A recursive divide-and-conquer approach for sparse principal component analysis}

\author{Qian~Zhao}
\ead{zhao.qian@stu.xjtu.edu.cn}

\author{Deyu~Meng\corref{cor1}}
\ead{dymeng@mail.xjtu.edu.cn}

\author{Zongben~Xu\corref{}}
\ead{zbxu@mail.xjtu.edu.cn}

\cortext[cor1]{Corresponding author. Tel.:~+86 13032904180;~fax:~+86 2982668559.}

\address{Institute for Information and System Sciences, School of Mathematics and Statistics, Xi'an Jiaotong University, Xi'an 710049, PR China}

\begin{abstract}
In this paper, a new method is proposed for sparse PCA based on the recursive divide-and-conquer methodology. The main idea is to separate the original sparse PCA problem into a series of much simpler sub-problems, each having a closed-form solution. By recursively solving these sub-problems in an analytical way, an efficient algorithm is constructed to solve the sparse PCA problem. The algorithm only involves simple computations and is thus easy to implement. The proposed method can also be very easily extended to other sparse PCA problems with certain constraints, such as the nonnegative sparse PCA problem. Furthermore, we have shown that the proposed algorithm converges to a stationary point of the problem, and its computational complexity is approximately linear in both data size and dimensionality. The effectiveness of the proposed method is substantiated by extensive experiments implemented on a series of synthetic and real data in both reconstruction-error-minimization and data-variance-maximization viewpoints.
\end{abstract}

\end{frontmatter}

\section{Introduction}

Principal component analysis (PCA) is one of the most classical
and popular tools for data analysis and dimensionality reduction, and has
a wide range of successful applications throughout science and engineering \cite{PCAbook}.
By seeking the so-called principal components (PCs), along which the data variance
is maximally preserved, PCA can always capture the intrinsic latent structure underlying data.
Such information greatly facilitates many further data processing tasks,
such as feature extraction and pattern recognition.

Despite its many advantages, the conventional PCA suffers from the
fact that each component is generally a linear combination of all
data variables, and all weights in the linear combination,
also called loadings, are typically non-zeros. In many
applications, however, the original variables have meaningful
physical interpretations. In biology, for example, each
variable of gene expression data corresponds to a certain gene.
In these cases, the derived PC loadings are always expected to
be sparse (i.e. contain fewer non-zeros) so as to facilitate their
interpretability. Moreover, in certain applications,
such as financial asset trading, the sparsity of the PC loadings is
especially expected since fewer nonzero loadings imply fewer
transaction costs.

Accordingly, sparse PCA has attracted much attention in the recent decade,
and a variety of methods for this topic have been developed \cite{Rotation,ST,SCoTLASS, SPCA, DSPCA, rSVD, GPower, EMPCA, SPPCA, DMP-SPCA, SPP, DC-PCA_C, DC-PCA_J, ALSPCA, PathSPCA, GSPCA, OPT_SPCA, PMD, ESPCA, Large-Scale, RSPCA, IE-SPCA}.
The first attempt for this topic is to make certain post-processing transformation, e.g.
rotation \cite{Rotation} by Jolliffe and simple thresholding \cite{ST} by Cadima and Jolliffe, on the PC loadings obtained by the conventional PCA to enforce sparsity.
Jolliffe and Uddin further advanced a SCoTLASS algorithm by simultaneously calculating
sparse PCs on the PCA model with additional $l_1$-norm penalty on loading vectors \cite{SCoTLASS}. Better results have been achieved by the SPCA
algorithm of Zou et al., which was developed based on iterative
elastic net regression \cite{SPCA}. D'Aspremont et al. proposed a
method, called DSPCA, for finding sparse PCs by solving a sequence
of semidefinite programming (SDP) relaxations of sparse PCA \cite{DSPCA}. Shen and Huang developed a
series of methods called sPCA-rSVD (including sPCA-rSVD$_{l_0}$, sPCA-rSVD$_{l_1}$, sPCA-rSVD$_{SCAD}$), computing
sparse PCs by low-rank matrix factorization under multiple sparsity-including penalties \cite{rSVD}.
Journ\'{e}e et al. designed four algorithms, denoted as GPower$_{l_0}$, GPower$_{l_1}$,
GPower$_{l_0,m}$, and GPower$_{l_1,m}$, respectively, for sparse PCA
by formulating the issue as non-concave maximization problems with
$l_0$- or $l_1$-norm sparsity-inducing penalties and extracting single
unit sparse PC sequentially or block units ones simultaneously \cite{GPower}. Based on probabilistic
generative model of PCA, some methods have also been attained \cite{EMPCA, SPP, SPPCA, DMP-SPCA}, e.g. the EMPCA method derived by
Sigg and Buhmann for sparse
and/or nonnegative sparse PCA \cite{EMPCA}.
Sriperumbudur et al. provided an iterative algorithm called DCPCA,
where each iteration consists of solving a quadratic programming
(QP) problem \cite{DC-PCA_C, DC-PCA_J}. Recently, Lu and Zhang
developed an augmented Lagrangian method (ALSPCA briefly) for sparse
PCA by solving a class of non-smooth constrained optimization
problems \cite{ALSPCA}. Additionally, d'Aspremont derived a PathSPCA algorithm
that computes a full set of solutions for all target numbers of nonzero coefficients \cite{PathSPCA}.

There are mainly two methodologies utilized by the current research on sparse PCA problem.
The first is the greedy approach, including DSPCA \cite{DSPCA}, sPCA-rSVD \cite{rSVD}, EMPCA \cite{EMPCA}, PathSPCA \cite{PathSPCA},
etc. These methods mainly focus on the
solving of one-sparse-PC model, and more sparse PCs can be
sequentially calculated on the deflated data matrix or data covariance \cite{Deflation}.
Under this methodology, the first several sparse PCs underlying the data
can generally be properly extracted, while the computation for more sparse PCs tends to be
incrementally invalidated due to the cumulation of computational error.
The second is the block approach. Typical methods include SCoTLASS \cite{SCoTLASS}, GPower$_{l_0,m}$, GPower$_{l_1,m}$ \cite{GPower}, ALSPCA \cite{ALSPCA}, etc. These methods aim to calculate
multiple sparse PCs at once by
utilizing certain block optimization techniques. The block approach for sparse PCA is expected to be more efficient than the greedy one to simultaneously attain multiple PCs, while is generally difficult to
elaborately rectify each individual sparse PC based on some specific requirements in practice
(e.g. the number of nonzero elements in each PC).

In this paper, a new methodology, called the recursive divide-and-conquer (ReDaC briefly),
is employed for solving the sparse PCA problem.
The main idea is to decompose the original large and complex problem of sparse PCA into a series of
small and simple sub-problems, and then recursively solve them.
Each of these sub-problems has a closed-form solution, which makes the new method simple
and very easy to implement.
On one hand, as compared with the greedy approach, the new method is expected
to integratively achieve a collection of appropriate sparse PCs of the problem
by iteratively rectifying each sparse PC in a recursive way. The group of sparse PCs attained by the proposed method
is further proved being a stationary solution of the original sparse PCA problem.
On the other hand,
as compared with the block approach, the new method can easily handle the constraints
superimposed on each individual sparse PC, such as certain sparsity and/or nonnegative constraints.
Besides, the computational complexity of the proposed method
is approximately linear in both data size and dimensionality, which makes it well-suited
to handle large-scale problems of sparse PCA.

In what follows, the main idea and the implementation details of the proposed method
are first introduced in Section 2. Its convergence and computational complexity are also
analyzed in this section.
The effectiveness of the proposed method is comprehensively substantiated
based on a series of empirical studies in Section 3.
Then the paper is concluded with a summary and outlook for future research.
Throughout the paper, we denote matrices, vectors and scalars by the upper-case bold-faced letters, lower-case bold-faced letters, and lower-case letters, respectively.

\section{The recursive divide-and-conquer method for sparse PCA}

In the following, we first introduce the fundamental models for the sparse PCA problem.

\subsection{Basic models of sparse PCA}

Denote the input data matrix as $\mathbf{X}=[\mathbf{x}_1,\mathbf{x}_2,\dots,\mathbf{x}_n]^T\in \mathbb{R}^{n\times
d}$, where $n$ and $d$ are the size and the dimensionality of the
given data, respectively. After a location transformation, we can
assume all $\{\mathbf{x}_i\}_{i=1}^{n}$ to have zero mean.
Let $\mathbf{\Sigma} = \frac{1}{n}\mathbf{X}^T\mathbf{X} \in \mathbb{R}^{d\times d}$ be the data covariance matrix.

The classical PCA can be solved through two types of optimization models \cite{PCAbook}.
The first is constructed by finding the $r(\leq d)$-dimensional linear
subspace where the variance of the input data $\mathbf{X}$ is maximized \cite{PCAvar}.
On this data-variance-maximization viewpoint, the PCA is formulated as the following
optimization model:
\begin{equation}\label{pca_var}
  \underset{\mathbf{V}}{\max}~\mathrm{Tr}(\mathbf{V}^T\mathbf{\Sigma}\mathbf{V})
    ~~~~s.t.~~\mathbf{V}^T\mathbf{V}=\mathbf{I},
\end{equation}
where $\mathrm{Tr}(\mathbf{A})$ denotes the trace of the matrix $\mathbf{A}$ and
$\mathbf{V}=(\mathbf{v}_1,\mathbf{v}_2,\dots,\mathbf{v}_r)\in \mathbb{R}^{d\times r}$ denotes the
array of PC loading vectors.
The second is formulated by seeking the $r$-dimensional linear subspace
on which the projected data and the original ones are as close as possible \cite{PCArec}.
On this reconstruction-error-minimization viewpoint, the PCA corresponds to the
following model:
\begin{equation}\label{pca_rec}
  \underset{\mathbf{U},\mathbf{V}}{\min}~
  \left\|\mathbf{X}-\mathbf{U}\mathbf{V}^T\right\|_F^2~~~~s.t.~~
 \mathbf{V}^T\mathbf{V}=\mathbf{I},
\end{equation}
where $\left\Vert \textbf{A}
\right\Vert _{F}$ is the Frobenius norm of $\textbf{A}$,
$\mathbf{V}\in \mathbb{R}^{d\times r}$ is the
matrix of PC loading array and $\mathbf{U}=(\mathbf{u}_1,\mathbf{u}_2,\dots,\mathbf{u}_r)\in \mathbb{R}^{n\times r}$
is the matrix of projected data. The two models
are intrinsically equivalent and can attain the same PC loading vectors \cite{PCAbook}.

Corresponding to the PCA models (\ref{pca_var}) and (\ref{pca_rec}), the sparse PCA problem
has the following two mathematical formulations\footnote{It should be noted that the orthogonality
constraints of PC loadings in (\ref{pca_var}) and (\ref{pca_rec})
are not imposed in (\ref{spca_var}) and (\ref{spca_rec}).
This is because simultaneously enforcing sparsity and orthogonality is generally
a very difficult (and perhaps unnecessary) task. Like most of the existing sparse PCA methods \cite{SPCA, DSPCA, rSVD, GPower}, we do not enforce orthogonal PCs in the models.
}:
\begin{equation}\label{spca_var}
  \underset{\mathbf{V}}{\max}~\mathrm{Tr}(\mathbf{V}^T\mathbf{\Sigma}\mathbf{V})
    ~~~~s.t.~~\mathbf{v}_i^T\mathbf{v}_i=1,~~~\|\mathbf{v}_i\|_p\leq t_i~(i=1,2,\dots,r),
\end{equation}
and
\begin{equation}\label{spca_rec}
  \underset{\mathbf{U},\mathbf{V}}{\min}~
  \left\|\mathbf{X}-\mathbf{U}\mathbf{V}^T\right\|_F^2~~~~s.t.~~\mathbf{v}_i^T\mathbf{v}_i=1,
  ~~~\|\mathbf{v}_i\|_p\leq t_i~(i=1,2,\dots,r),
\end{equation}
where $p=0$ or $1$ and the corresponding $\|\mathbf{v}\|_p$ denotes the $l_0$- or the $l_1$-norm
of $\mathbf{v}$, respectively. Note that the involved $l_0$ or $l_1$ penalty in the above models
(\ref{spca_var}) and (\ref{spca_rec}) tends to enforce sparsity of the output PCs.
Methods constructed on (\ref{spca_var}) include SCoTLASS \cite{SCoTLASS}, DSPCA \cite{DSPCA}, DCPCA \cite{DC-PCA_C, DC-PCA_J}, ALSPCA \cite{ALSPCA}, etc., and those related to (\ref{spca_rec})
include SPCA \cite{SPCA}, sPCA-rSVD \cite{rSVD}, SPC \cite{PMD}, GPower \cite{GPower}, etc.
In this paper, we will construct our method on the reconstruction-error-minimization model (\ref{spca_rec}),
while our experiments will verify that the proposed method also performs well
based on the data-variance-maximization criterion.

\subsection{Decompose original problem into small and simple sub-problems}

The objective function of the sparse PCA model (\ref{spca_rec}) can be equivalently formulated as follows:
$$
\left\Vert \mathbf{X}-\mathbf{UV}^{T}\right\Vert _{F}^2=\left\Vert \mathbf{%
X}-{\sum\nolimits }_{j=1}^{r}\mathbf{u}_{j}\mathbf{v}_{j}^{T}\right\Vert
_{F}^2=\left\Vert \mathbf{E}_{i}-\mathbf{u}_{i}\mathbf{v}%
_{i}^{T}\right\Vert _{F}^2,
$$
where $\textbf{E}_{i}=\textbf{X}-{\sum}_{j\neq i} \mathbf{u}_{j}\mathbf{v}_{j}^{T}$.
It is then easy to separate the original large
minimization problem, which is with respect to $\textbf{U}$
and $\textbf{V}$, into a series of small minimization problems, which are
each with respect to a column vector $\mathbf{u}_{i}$ of $\mathbf{U}$ and $\mathbf{v}_{i}$
of $\mathbf{V}$ for $i=1,2,\dots,r$, respectively, as follows:
\begin{equation}\label{sub_vk}
\underset{\mathbf{v}_i}{\min}\left\|\mathbf{E}_i-\mathbf{u}_i\mathbf{v}_i^T\right\|_F^2
~~~~s.t.~~\mathbf{v}_i^T\mathbf{v}_i=1,
  ~~~\|\mathbf{v}_i\|_p\leq t_i,
\end{equation}
and
\begin{equation}\label{sub_uk}
\underset{\mathbf{u}_i}{\min}\left\|\mathbf{E}_i-\mathbf{u}_i\mathbf{v}_i^T\right\|_F^2.
\end{equation}
Through recursively optimizing these small sub-problems, the recursive divide-and-conquer (ReDaC) method
for solving the sparse PCA model (\ref{spca_rec}) can then be naturally constructed.

It is very fortunate that both the minimization problems in (\ref{sub_vk}) and (\ref{sub_uk}) have closed-form solutions. This implies
that the to-be-constructed ReDaC method can be fast and efficient, as presented in
the following sub-sections.

\subsection{The closed-form solutions of (\ref{sub_vk}) and (\ref{sub_uk})}

For the convenience of denotation, we first rewrite (\ref{sub_vk}) and (\ref{sub_uk}) as the following forms:
\begin{equation}
\underset{\mathbf{v}}{\min }~~\left\Vert \mathbf{E}-\mathbf{uv}^{T}\right\Vert
_{F}^{2}~~~~s.t.~~\mathbf{v}^{T}\mathbf{v}=1,~~~\Vert \mathbf{v}\Vert
_{p}\leq t,   \label{ev}
\end{equation}
and%
\begin{equation}
\underset{\mathbf{u}}{\min }~~\left\Vert \mathbf{E}-\mathbf{uv}^{T}\right\Vert
_{F}^{2},   \label{eu}
\end{equation}
where $\mathbf{u}$ is $n$-dimensional and $\mathbf{v}$ is $d$-dimensional.
Since the objective function $\left\Vert \mathbf{E}-\mathbf{uv}^{T}\right\Vert
_{F}^{2}$ can be equivalently transformed
as:
\begin{equation*}
\begin{split}
\left\Vert \mathbf{E}-\mathbf{u}\mathbf{v}^{T}\right\Vert _{F}^{2}& =\mathrm{%
Tr}((\mathbf{E}-\mathbf{u}\mathbf{v}^{T})^{T}(\mathbf{E}-\mathbf{u}\mathbf{v}%
^{T})) \\
& =\Vert \mathbf{E}\Vert _{F}^{2}-2\mathrm{Tr}(\mathbf{E}^{T}\mathbf{u}%
\mathbf{v}^{T})+\mathrm{Tr}(\mathbf{v}\mathbf{u}^{T}\mathbf{u}\mathbf{v}^{T})
\\
& =\Vert \mathbf{E}\Vert _{F}^{2}-2\mathbf{u}^{T}\mathbf{E}\mathbf{v}+%
\mathbf{u}^{T}\mathbf{u}\mathbf{v}^{T}\mathbf{v},
\end{split}%
\end{equation*}%
(\ref{ev}) and (\ref{eu}) are equivalent to the following
optimization problems, respectively:%
\begin{equation}
\underset{\mathbf{v}}{\max }~~(\mathbf{E}^{T}\mathbf{u})^{T}\mathbf{v}%
~~~~s.t.~~\mathbf{v}^{T}\mathbf{v}=1,~~~\Vert \mathbf{v}\Vert _{p}\leq t, \label{evnew}
\end{equation}%
and
\begin{equation}
\underset{\mathbf{u}}{\min }~~\mathbf{u}^{T}\mathbf{u}-2(\mathbf{Ev})^{T}\mathbf{u}. \label{eunew}
\end{equation}
The closed-form solutions of (\ref{evnew}) and (\ref{eunew}), i.e. (\ref{ev}) and (\ref{eu}), can then be presented as follows.

We present the closed-form solution to (\ref{eu}) in the following theorem.

\begin{theorem}
The optimal solution of (\ref{eu}) is $\mathbf{u}^{\ast }(\mathbf{v})=\mathbf{Ev}$.
\end{theorem}

The theorem is very easy to prove by calculating where the gradient of $\mathbf{u}^{T}\mathbf{u}-2(\mathbf{Ev})^{T}\mathbf{u}$ is equal to zero.
We thus omit the proof.

In the $p=0$ case, the closed-form solution to (\ref{evnew}) is presented in the following
theorem. Here, we denote $\mathbf{w}=\mathbf{E}^{T}\mathbf{u}$, and $%
hard_{\lambda }(\mathbf{w})$ the hard thresholding function, whose $i$-th
element corresponds to $I(|w_{i}|\geq \lambda )w_{i}$, where $w_{i}$
is the $i$-th element of $\mathbf{w}$\ and $I(x)$ (equals $1$ if $x$ is ture%
, and $0$ otherwise)\ is the indicator function. The proof of the theorem is provided in Appendix A.

\begin{theorem}
The optimal solution of
\begin{equation}
\underset{\mathbf{v}}{\max }~~\mathbf{w}^{T}\mathbf{v}~~~~s.t.~~\mathbf{v}^{T}%
\mathbf{v}=1, ~~~\Vert \mathbf{v}\Vert _{0}\leq t,   \label{tv0}
\end{equation}%
is given by:%
\begin{equation*}
\mathbf{v}^{\ast }_0(\mathbf{w},t)=\left\{
\begin{array}{cl}
\phi , & t<1\text{,} \\
\frac{hard_{\theta_k}(\mathbf{w})}{\Vert hard_{\theta_k}(\mathbf{w})\Vert _{2}}\mathbf{,}
& k\leq t<k+1~~(k=1,2,\dots ,d-1), \\
\frac{\mathbf{w}}{\Vert \mathbf{w}\Vert _{2}} & t\geq d\text{,}%
\end{array}%
\right.
\end{equation*}
where $\theta_k$ denotes the $k$-th largest element of $|\mathbf{w}|$.
\end{theorem}

In the above theorem, $\phi$ denotes the empty set, implying that when $t<1$,
the optimum of (\ref{tv0}) does not exist.

In the $p=1$ case, (\ref{ev}) has the following closed-form solution.
In the theorem, we denote $f_{\mathbf{w}}\mathbf{(\lambda )}=%
\frac{soft_{\lambda }(\mathbf{w})}{\left\Vert soft_{\lambda }(\mathbf{w}%
)\right\Vert _{2}}$, where $soft_{\lambda }(\mathbf{w})$ represents the soft
thresholding function $sign(\mathbf{w})(|\mathbf{w|}-\lambda )_{+}$, where
$(\mathbf{x})_{+}$ represents the vector attained by projecting $\mathbf{x}$ to its nonnegative orthant,
and $(I_{1},I_{2},\dots ,I_{d})$ denotes the permutation of $(1,2,\dots ,d)$ based on
the ascending order of $|\mathbf{w|}=(\mathbf{|}w_{1}\mathbf{|},\mathbf{|}%
w_{2}\mathbf{|},\dots ,\mathbf{|}w_{d}\mathbf{|})^{T}$.

\begin{theorem}
The optimal solution of
\begin{equation} \label{tu0}
\underset{\mathbf{v}}{\max }~~\mathbf{w}^{T}\mathbf{v}~~~~s.t.~~\mathbf{v}^{T}%
\mathbf{v}=1,~~~\Vert\mathbf{v}\Vert _{1}\leq t,
\end{equation}%
is given by:%
\begin{equation*}
\mathbf{v}^{\ast }_1(\mathbf{w},t)=\left\{
\begin{array}{cl}
\phi , & t<1, \\
f_{\mathbf{w}}\mathbf{(\lambda }_{k}\mathbf{),} & \|f_{\mathbf{w}}%
\mathbf{(}|w_{I_{k}}|)\|_1\leq t< \|f_{\mathbf{w}}\mathbf{(}|w_{I_{k-1}}|)\|_1~~(k=2,3,\dots
,d-1), \\
f_{\mathbf{w}}\mathbf{(\lambda }_{1}\mathbf{),} & \|f_{\mathbf{w}}%
\mathbf{(}|w_{I_{1}}|)\|_1\leq t<\sqrt{d},  \\
f_{\mathbf{w}}(0), & t\geq \sqrt{d},%
\end{array}%
\right.
\end{equation*}%
where for $k=1,2,\dots,d-1$,
\begin{equation*}
\mathbf{\lambda }_{k}=\frac{(m-t^{2})(\sum_{i=1}^{m}a_{i})-\sqrt{%
t^{2}(m-t^{2})(m\sum_{i=1}^{m}a_{i}^{2}-(\sum_{i=1}^{m}a_{i})^{2})}}{m(m-t^{2})}\text{,}
\end{equation*}%
where $(a_{1},a_{2},\dots ,a_{m})=(|w_{I_{k}}|,|w_{I_{k+1}}|,\dots
,|w_{I_{d}}|)$, $m=d-k+1$.
\end{theorem}

It should be noted that we have proved that $\|f_{\mathbf{w}}\mathbf{(}|w_{I_{d-1}}|)\|_1=1$ and $\|f_{%
\mathbf{w}}\mathbf{(\lambda )}\|_1$ is a monotonically decreasing function with
respect to $\mathbf{\lambda }$
in Lemma 1 of the appendix. This means that we can conduct the optimum $\mathbf{v}^{\ast }(\mathbf{w})$
of the optimization problem (\ref{ev}) for any $\mathbf{w}$ based on the above
theorem.

The ReDaC algorithm can then be easily constructed based on Theorems 1-3.

\subsection{The recursive divide-and-conquer algorithm for sparse PCA}

The main idea of the new algorithm is to recursively optimize each column, $\mathbf{u}_{i}$ of $\mathbf{U}$ or $\mathbf{v}_{i}$ of $\mathbf{V}$ for $i=1,2,\dots,r$, with other $\mathbf{u}_{j}$s and
$\mathbf{v}_{j}$s ($j\neq i$) fixed. The process is summarized as follows:
\begin{itemize}
  \item Update each column $\mathbf{v}_{i}$ of $\mathbf{V}$ for $i=1,2,\dots,r$ by the closed-form solution
  of (\ref{sub_vk}) attained from Theorem 2 (for $p=0$) or Theorem 3 (for $p=1$).
  \item Update each column $\mathbf{u}_{i}$ of $\mathbf{U}$ for $i=1,2,\dots,r$ by the closed-form solution
  of (\ref{sub_uk}) calculated from Theorem 1.
\end{itemize}
Through implementing the above procedures iteratively, $\textbf{U}$ and $\textbf{V}$ can be recursively updated until the stopping criterion is satisfied. We summarize the aforementioned ReDaC technique as Algorithm \ref{algo_spca}.

\begin{algorithm} [t]
\caption{ReDaC algorithm for sparse PCA}\label{algo_spca}
\begin{algorithmic}[1]
\Require Data matrix $\mathbf{X}\in R^{n\times d}$, number of sparse PCs $r$, sparsity parameters $\mathbf{t}=(t_1,\dots,t_r)$.
\State Initialize $\mathbf{U}=(\mathbf{u}_1,\mathbf{u}_2,\dots,\mathbf{u}_r)\in R^{n\times r}$, $\mathbf{V}=(\mathbf{v}_1,\mathbf{u}_2,\dots,\mathbf{v}_r)\in R^{d\times r}$.
\Repeat
  \For{$i=1,\dots r$}
    \State Compute $\mathbf{E}_i=\mathbf{X}-\sum_{j\neq i}\mathbf{u}_j\mathbf{v}_j^T$.
    \State Update $\mathbf{v}_i$ via solving (\ref{sub_vk}) based on Theorem 2 (for $p=0$) or Theorem 3 (for $p=1$).
    \State Update $\mathbf{u}_i$ via solving (\ref{sub_uk}) based on Theorem 1.
  \EndFor
\Until stopping criterion satisfied.
\Ensure The sparse PC loading vectors $\mathbf{V}=(\mathbf{v}_1,\mathbf{v}_2,\dots,\mathbf{v}_r)$.
\end{algorithmic}
\end{algorithm}

We then briefly discuss how to specify the stopping criterion of the algorithm.
The objective function of the sparse PCA model (\ref{spca_rec})
is monotonically decreasing in the iterative process of Algorithm 1 since
each of the step 5 and step 6 in the iterations makes an exact optimization for a column
vector $\mathbf{u}_i$ of $\mathbf{U}$ or $\mathbf{v}_i$ of $\mathbf{V}$, with all of the others fixed.
We can thus terminate the iterations of the algorithm
when the updating rate of $\mathbf{U}$ or $\mathbf{V}$ is smaller than some preset threshold,
or the maximum number of iterations is reached.

Now we briefly analyze the computational complexity of the proposed ReDaC algorithm.
It is evident that the computational complexity of Algorithm \ref{algo_spca}
is essentially determined by the iterations between step 5 and step 6, i.e. the calculation of the closed-form solutions of $\mathbf{v}_i$ and $\mathbf{u}_i$ of $\mathbf{V}$ and $\mathbf{U}$, respectively. To compute $\mathbf{u}_i$, only simple operations are involved and the computation needs $O(nd)$ cost. To compute $\mathbf{v}_i$, a sorting for the elements of the $d$-dimensional vector $|\mathbf{w}|=|\mathbf{E}^{T}\mathbf{u}|$ is
required, and the total computational cost is around
$O(nd\log d)$ by applying the well-known heap
sorting algorithm \cite{heap}. The whole process of the algorithm thus requires around $O(rnd\log d)$ computational cost in each iteration. That is, the computational complexity of the proposed algorithm is approximately linear in both the size and the dimensionality of input data.

\subsection{Convergence analysis}

In this section we evaluate the convergence of the proposed algorithm.

The convergence of our algorithm can actually be implied by the monotonic decrease of the cost function
of (\ref{spca_rec}) during the iterations of the algorithm. In specific, in each iteration of the algorithm,
step 5 and step 6 optimize the column
vector $\mathbf{u}_i$ of $\mathbf{U}$ or $\mathbf{v}_i$ of $\mathbf{V}$, with all of the others fixed, respectively.
Since the objective function of (\ref{spca_rec}) is evidently lower bounded ($\geq0$), the algorithm is guaranteed to be convergent.

We want to go a further step to evaluate where the algorithm converges.
Based on the formulation of
the optimization problem (\ref{spca_rec}), we can construct a specific function as follows:
\begin{equation}
f(\mathbf{u}_{1},\dots,\mathbf{u}_{r},\mathbf{v}_{1},\dots,\mathbf{v}_{r})
=f_0(\mathbf{u}_{1},\dots,\mathbf{u}_{r},\mathbf{v}_{1},\dots,\mathbf{v}_{r})+\sum_{i=1}^rf_{i}(\mathbf{v}_i).
\end{equation}
where
\begin{equation*}
f_0(\mathbf{u}_{1},\dots,\mathbf{u}_{r},\mathbf{v}_{1},\dots,\mathbf{v}_{r})=\left\|\mathbf{X}-\mathbf{U}\mathbf{V}^T\right\|_F^2
=\left\|\mathbf{X}-\sum\nolimits_{i=1}^r\mathbf{u}_i\mathbf{v}_i^T\right\|_F^2,
\end{equation*}
and for each of $i=1,\dots,r$, $f_{i}(\mathbf{v}_i)$ is an indicator function defined as:
\begin{equation*}
f_{i}(\mathbf{v}_i)=
  \begin{cases}
    0, & \textrm{if}~~\|\mathbf{v}_i\|_p\leq t_i~~\textrm{and}~~\mathbf{v}_i^T\mathbf{v}_i=1, \\
    \infty, & \textrm{otherwise}.
  \end{cases}
\end{equation*}
It is then easy to show that the constrained optimization problem (\ref{spca_rec}) is equivalent to the unconstrained problem
\begin{equation} \label{Block}
\underset{\{\mathbf{u}_i,\mathbf{v}_i\}_{i=1}^r}{\min}f(\mathbf{u}_{1},\dots,\mathbf{u}_{r},\mathbf{v}_{1},\dots,\mathbf{v}_{r}).
\end{equation}
The proposed ReDaC algorithm can then be viewed as a block coordinate descent (BCD) method for solving (\ref{Block}) \cite{BCD-Convergence},
by alteratively optimizing $\mathbf{u}_{i},\mathbf{v}_{i}$, $i=1,2,\dots,r$, respectively. Then the following theorem
implies that our algorithm can converge to a stationary point of the problem.

\begin{theorem}[\cite{BCD-Convergence}]
Assume that the level set $X^0=\{x:f(x)\leq f(x^0)\}$ is compact and that $f$ is continuous on $X^0$. If
$f(\mathbf{u}_{1},\dots,\mathbf{u}_{r},\mathbf{v}_{1},\dots,\mathbf{v}_{r})$ is regular and has at most one minimum in each $\mathbf{u}_{i}$ and
$\mathbf{v}_{i}$ with others fixed for $i=1,2,\dots,r$, then the sequence
$(\mathbf{u}_{1},\dots,\mathbf{u}_{r},\mathbf{v}_{1},\dots,\mathbf{v}_{r})$ generated by Algorithm 1
converges to a stationary point of $f$.
\end{theorem}

In the above theorem, the assumption that the function $f$, as defined in (\ref{Block}), is regular holds under the condition
that
$dom(f_0)$ is open and $f_0$ is Gateaux-differentiable on $dom(f_0)$ (Lemma 3.1 under Condition A1 in \cite{BCD-Convergence}).
Based on Theorems 1-3, we can also easily see that $f(\mathbf{u}_{1},\dots,\mathbf{u}_{r},\mathbf{v}_{1},\dots,\mathbf{v}_{r})$ has unique minimum in each $\mathbf{u}_{i}$ and $\mathbf{v}_{i}$ with others fixed. The above theorem can then be naturally followed by Theorem 4.1(c) in \cite{BCD-Convergence}.

Another advantage of the proposed ReDaC methodology is that it can be easily extended
to other sparse PCA applications when certain constraints are needed for output sparse PCs.
In the following section we give one of the extensions of our methodology --- nonnegative sparse PCA problem.

\subsection{The ReDaC method for nonnegative sparse PCA}

The nonnegative sparse PCA \cite{NSPCA} problem differs from the conventional sparse PCA in its nonnegativity constraint
imposed on the output sparse PCs. The nonnegativity property of this problem is especially important in some applications such as
microeconomics, environmental science, biology, etc. \cite{NN}. The corresponding optimization model is written as
follows:
\begin{equation}\label{nspca}
  \underset{\mathbf{U},\mathbf{V}}{\min}~
  \left\|\mathbf{X}-\mathbf{U}\mathbf{V}^T\right\|_F^2~~~~s.t.~~\mathbf{v}_i^T\mathbf{v}_i=1,
  ~~~\|\mathbf{v}_i\|_p\leq t_i, ~~\mathbf{v}_i\succeq 0~(i=1,2,\dots,r),
\end{equation}
where $\mathbf{v}_i\succeq 0$ means that each element of $\mathbf{v}_i$ is greater than or equal to 0.

By utilizing the similar recursive divide-and-conquer strategy, this problem can be
separated into a series of small minimization problems, each with respect to a column vector
$\mathbf{u}_{i}$ of $\mathbf{U}$ and $\mathbf{v}_{i}$ of $\mathbf{V}$ for $i=1,2,\dots,r$, respectively, as follows:
\begin{equation}\label{ns_vk}
\underset{\mathbf{v}_i}{\min}\left\|\mathbf{E}_i-\mathbf{u}_i\mathbf{v}_i^T\right\|_F^2
~~~~s.t.~~\mathbf{v}_i^T\mathbf{v}_i=1,
  ~~~\|\mathbf{v}_i\|_p\leq t_i,~~~\mathbf{v}_i\succeq 0
\end{equation}
and
\begin{equation}\label{ns_uk}
\underset{\mathbf{u}_i}{\min}\left\|\mathbf{E}_i-\mathbf{u}_i\mathbf{v}_i^T\right\|_F^2,
\end{equation}
where $p=0$ or $1$.
Since (\ref{ns_uk}) is of the same formulation as (\ref{sub_uk}), we only need to discuss how to
solve (\ref{ns_vk}). For the convenience of denotation, we first rewrite (\ref{ns_vk}) as:
\begin{equation}
\underset{\mathbf{v}}{\min }~~\left\Vert \mathbf{E}-\mathbf{uv}^{T}\right\Vert
_{F}^{2}~~~~s.t.~~\mathbf{v}^{T}\mathbf{v}=1,~~~\Vert \mathbf{v}\Vert
_{p}\leq t,~~~\mathbf{v}\succeq 0.   \label{nsv}
\end{equation}
The closed-form solution of (\ref{nsv}) is given in the following theorem.

\begin{theorem}
The closed-form solution of (\ref{nsv}) is $\mathbf{v}%
_{p}^{\ast }((\mathbf{w})_{+},t)$ ($p=0,1$), where $\mathbf{w=E}^{T}\mathbf{u}$,
and $\ \mathbf{v}_{0}^{\ast }(\cdot, \cdot)$ and $\mathbf{v}_{1}^{\ast }(\cdot, \cdot )$\
are defined in Theorem 2 and Theorem 3, respectively.
\end{theorem}

By virtue of the closed-form solution of (\ref{nsv}) given by Theorem 5, we can now construct the ReDaC algorithm
for solving nonnegative sparse PCA model (\ref{nspca}). Since the algorithm differs from Algorithm 1 only in step 5
(i.e. updating of $\mathbf{v}_{i}$), we only list this step in Algorithm 2.

\begin{algorithm}[t]
\caption{ReDaC algorithm for nonnegative sparse PCA}\label{algo_nspca}
\begin{algorithmic}[0]
\State \footnotesize 5: \normalsize ~~~Update $\mathbf{v}_i$ via solving (\ref{ns_vk}) based on Theorem 5.
\end{algorithmic}
\end{algorithm}

We then substantiate the effectiveness of the proposed ReDaC algorithms for sparse PCA and nonnegative sparse
PCA through experiments in the next section.

\section{Experiments}

To evaluate the performance of the proposed ReDaC algorithm on the sparse PCA problem, we
conduct experiments on a series of synthetic and real data sets. All the experiments are implemented on Matlab 7.11(R2010b) platform in a PC with
AMD Athlon(TM) 64 X2 Dual 5000+@2.60 GHz (CPU), 2GB (memory), and Windows XP (OS). In all experiments, the SVD method is utilized for initialization.
The proposed algorithm under both $p=0$ and $p=1$ was implemented in all experiments and mostly have a similar performance. We thus
only list the better one throughout.

\subsection{Synthetic simulations}

Two synthetic data sets are first utilized to evaluate the performance of the proposed algorithm on recovering
the ground-truth sparse principal components underlying data.

\subsubsection{Hastie data}

Hastie data set was first proposed by Zou et al. \cite{SPCA} to illustrate the advantage of sparse PCA
over conventional PCA on sparse PC extraction. So far this data set has
become one of the most frequently utilized benchmark data for testing the effectiveness of sparse PCA methods.
The data set is generated in the following way: first, three hidden factors $V_1$, $V_2$ and $V_3$ are created as:
\begin{equation*}
V_1\thicksim \mathcal{N}(0,290),~~V_2\thicksim \mathcal{N}(0,300),~~V_3=0.3V_1+0.925V_2+\varepsilon,
\end{equation*}
where $\varepsilon\thicksim \mathcal{N}(0,1)$, and $V_1$, $V_2$ and $\varepsilon$ are independent; afterwards, $10$ observable variables are generated as:
\begin{equation*}
\begin{split}
&X_i=V_1+\varepsilon_i^1,~~i=1,2,3,4,\\
&X_i=V_2+\varepsilon_i^2,~~i=5,6,7,8,\\
&X_i=V_3+\varepsilon_i^3,~~i=9,10,\\
\end{split}
\end{equation*}
where $\varepsilon_i^j\thicksim \mathcal{N}(0,1)$ and all $\varepsilon_i^j$s are independent. The data so generated are of intrinsic sparse PCs \cite{SPCA}: the first recovers the factor $V_2$ only using $(X_5,X_6,X_7,X_8)$, and the second recovers $V_1$ only utilizing $(X_1,X_2,X_3,X_4)$.

We generate $100$ sets of data, each contains $1000$ data generated in the aforementioned way, and apply Algorithm 1 to them to extract the first
two sparse PCs. The results show that our algorithm can perform well in all experiments. In specific, the proposed ReDaC algorithm faithfully delivers
the ground-truth sparse PCs in all experiments. The effectiveness of the proposed algorithm is thus easily substantiated in this series of benchmark data.

\subsubsection{Synthetic toy data}\label{sythetic_spca}

As \cite{rSVD} and \cite{GPower}, we adopt another interesting toy data, with intrinsic sparse PCs, to evaluate
the performance of the proposed method. The data are generated from the Gaussian distribution
$\mathcal{N}(\mathbf{0},\mathbf{\Sigma})$ with mean $\mathbf{0}$ and covariance $\mathbf{\Sigma}\in \mathbb{R}^{10\times10}$,
which is calculated by
\begin{equation*}
  \mathbf{\Sigma}=\sum_{j=1}^{10}c_j\mathbf{v}_j\mathbf{v}_j^T.
\end{equation*}
Here, $(c_1,c_2,...,c_{10})$, the eigenvalues of the covariance matrix $\mathbf{\Sigma}$, are pre-specified as $(250, 240, 50, 50, 6, 5, 4, 3, 2, 1)$, respectively,
and $(\mathbf{v}_1,\mathbf{v}_2,...,\mathbf{v}_{10})$ are $10$-dimensional orthogonal vectors, formulated by
\begin{equation*}
\begin{split}
  &\mathbf{v}_1=(0.422, 0.422, 0.422, 0.422, 0, 0, 0, 0, 0.380, 0.380)^T,\\
  &\mathbf{v}_2=(0, 0, 0, 0, 0.489, 0.489, 0.489, 0.489, -0.147, 0.147)^T,
\end{split}
\end{equation*}
and the rest being generated by applying Gram-Schmidt orthonormalization to $8$ randomly valued
$10$-dimensional vectors. It is easy to see that the data generated under this distribution
are of first two sparse PC vectors $\mathbf{v}_1$ and $\mathbf{v}_2$.

Four series of experiments, each involving $1000$ sets of data generated from $\mathcal{N}(\mathbf{0},\mathbf{\Sigma})$,
are utilized, with sample sizes $500$, $1000$, $2000$, $5000$, respectively. For each experiment,
the first two PCs, $\hat{\mathbf{v}}_1$ and $\hat{\mathbf{v}}_2$, are calculated by a sparse PCA method and then
if both $|\hat{\mathbf{v}}_1^T\mathbf{v}_1|\geq0.99$ and $|\hat{\mathbf{v}}_2^T\mathbf{v}_2|\geq0.99$ are satisfied,
the method is considered as a success. The proposed ReDaC method, together with
the conventional PCA and $12$ current sparse PCA methods, including SPCA \cite{SPCA}, DSPCA \cite{DSPCA}, PathSPCA \cite{PathSPCA}, sPCA-rSVD$_{l_0}$, sPCA-rSVD$_{l_1}$, sPCA-rSVD$_{SCAD}$ \cite{rSVD}, EMPCA \cite{EMPCA}, GPower$_{l_{0}}$, GPower$_{l_{1}}$, GPower$_{l_{0,m}}$, GPower$_{l_{1,m}}$ \cite{GPower} and
ALSPCA \cite{ALSPCA}, have been implemented, and the success times for four series of experiments
have been recorded and summarized, respectively. The results
are listed in Table \ref{result_simu_identify}.

\begin{table}[t]
\centering \caption{Comparison of success times of PCA and different sparse PCA methods in synthetic toy experiments with sample size varying. The best results are highlighted in bold.}
\label{result_simu_identify}
\vspace{0.8em}
\footnotesize
\begin{tabular}{ccccc}
\hline
   & $n=500$ & $n=1000$ & $n=2000$ & $n=5000$ \\
\hline
  PCA & $0$ & $0$ & $0$ & $0$ \\
  SPCA & $566$ & $673$ & $756$ & $839$ \\
  DSPCA & $211$ & $203$ & $138$ & $62$ \\
  PathSPCA & $189$ & $187$ & $186$ & $171$ \\
  sPCA-rSVD$_{l_0}$ & $646$ & $702$ & $797$ & $906$ \\
  sPCA-rSVD$_{l_1}$ & $649$ & $715$ & $806$ & $909$ \\
  sPCA-rSVD$_{\textrm{SCAD}}$ & $649$ & $715$ & $806$ & $909$ \\
  EMPCA & $649$ & $715$ & $806$ & $909$ \\
  GPower$_{l_0}$ & 155 & 154 & 155 & 139 \\
  GPower$_{l_1}$ & 122 & 127 & 126 & 126 \\
  GPower$_{l_{0,m}}$ & 91 & 76 & 71 & 16 \\
  GPower$_{l_{1,m}}$ & 90 & 92 & 88 & 82 \\
  ALSPCA & 669 & \textbf{749} & 826 & 927  \\
  ReDaC & $\textbf{676}$ & $748$ & $\textbf{827}$ & $\textbf{928}$ \\
\hline
\end{tabular}
\end{table}

The advantage of the proposed ReDaC algorithm can be easily observed from Table \ref{result_simu_identify}.
In specific, our method always attains the highest or second highest success times
(in the size $1000$ case, $1$ less than ALSPCA) as compared with the other utilized methods
in all of the four series of experiments. Considering that the ALSPCA method, which is the only
comparable method in these experiments, utilizes
strict constraints on the orthogonality of output PCs while the ReDaC method does not utilize any
prior ground-truth information of data, the capability of the proposed method
on sparse PCA calculation can be more prominently verified.

\subsection{Experiments on real data}

In this section, we further evaluate the performance of the proposed ReDaC method on two real data sets, including the pitprops and colon data.
Two quantitative criteria are employed for performance assessment. They are designed in the viewpoints of reconstruction-error-minimization
and data-variance-maximization, respectively, just corresponding to the original formulations (\ref{spca_rec}) and (\ref{spca_var}) for sparse PCA problem.
\begin{itemize}
  \item \emph{Reconstruction-error-minimization criterion: RRE.}
  Once sparse PC loading matrix $\mathbf{V}$ is obtained by a method, the input data can then be reconstructed by
  $\hat{\mathbf{X}}=\hat{\mathbf{U}}\mathbf{V}^T$, where $\hat{\mathbf{U}}=\mathbf{X}\mathbf{V}(\mathbf{V}^T\mathbf{V})^{-1}$, attained by
  the least square method. Then the relative reconstruction error (RRE) can be calculated by
      \begin{equation*}
        \textrm{RRE}=\frac{\|\mathbf{X}-\hat{\mathbf{X}}\|_F}{\|\mathbf{X}\|_F},
      \end{equation*}
      to assess the performance of the utilized method in data reconstruction point of view.
  \item \emph{Data-variance-maximization criterion: PEV.}
  After attaining the sparse PC loading matrix $\mathbf{V}$, the input data can then be reconstructed by $\hat{\mathbf{X}}=\mathbf{X}\mathbf{V}(\mathbf{V}^T\mathbf{V})^{-1}\mathbf{V}^T$, as aforementioned.
  And thus the variance of the reconstructed data can be computed by $\mathrm{Tr}(\frac{1}{n}\hat{\mathbf{X}}^T\hat{\mathbf{X}})$.
  The percentage of explained variance (PEV, \cite{rSVD}) of the reconstructed data from the original one can then be
  calculated by
      \begin{equation*}
        \textrm{PEV}=\frac{\mathrm{Tr}(\frac{1}{n}\hat{\mathbf{X}}^T\hat{\mathbf{X}})}{\mathrm{Tr}(\frac{1}{n}\mathbf{X}^T\mathbf{X})}\times100\%
        =\frac{\mathrm{Tr}(\hat{\mathbf{X}}^T\hat{\mathbf{X}})}{\mathrm{Tr}(\mathbf{X}^T\mathbf{X})}\times100\%,
      \end{equation*}
    to evaluate the performance of the utilized method in data variance point of view.
\end{itemize}

\subsubsection{Pitprops data}

\begin{figure}[t]
\begin{center}
\scalebox{0.48}{\includegraphics[viewport=90 60 963 427]{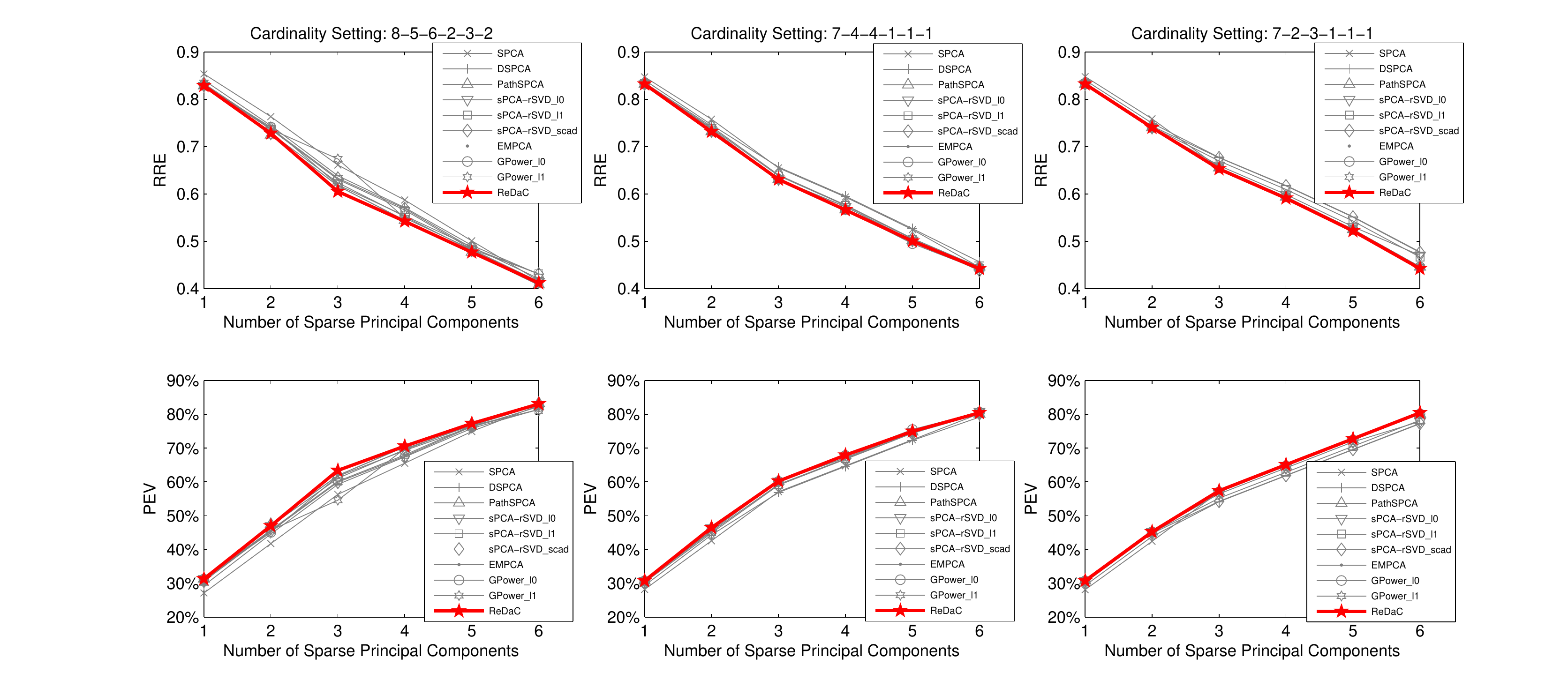}}
\end{center}
\caption{The tendency curves of RRE and PEV with respect to the number of extracted sparse PCs
attained by different sparse PCA methods
on pitprops data. Three cardinality settings for the extracted sparse PCs are utilized, including
8-5-6-2-3-2, 7-4-4-1-1-1 and 7-2-3-1-1-1.} \label{piptrop_curve}
\end{figure}

The pitprops data set, consisting of $180$ observations and $13$ measured variables, was first introduced by Jeffers \cite{Pitprops} to show the difficulty of interpreting PCs. This data set is one of the most commonly utilized examples for sparse PCA evaluation, and thus is also employed to
testify the effectiveness of the proposed ReDaC method. The comparison methods include SPCA \cite{SPCA}, DSPCA \cite{DSPCA}, PathSPCA \cite{PathSPCA}, sPCA-rSVD$_{l_0}$, sPCA-rSVD$_{l_1}$, sPCA-rSVD$_{SCAD}$ \cite{rSVD}, EMPCA \cite{EMPCA}, GPower$_{l_{0}}$, GPower$_{l_{1}}$, GPower$_{l_{0,m}}$, GPower$_{l_{1,m}}$ \cite{GPower} and
ALSPCA \cite{ALSPCA}. For each utilized method, $6$ sparse PCs are extracted from the pitprops data, with different cardinality settings:
8-5-6-2-3-2 (altogether $26$ nonzero elements), 7-4-4-1-1-1 (altogether $18$ nonzero elements, as set in \cite{SPCA}) and 7-2-3-1-1-1 (altogether $15$ nonzero elements, as set in \cite{DSPCA}), respectively. In each experiment, both the RRE and PEV values, as defined above, are calculated, and the results are summarized in Table \ref{result_pitprops}.
Figure \ref{piptrop_curve} further shows the the RRE and PEV curves attained by different sparse PCA methods in all experiments for more illumination. It should be noted that the GPower$_{l_0,m}$, GPower$_{l_1,m}$ and ALSPCA methods employ the block methodology, as introduced in the introduction of the paper, and calculate all sparse PCs at once while cannot sequentially derive different numbers of sparse PCs with preset cardinality settings. Thus the results of these methods reported in Table \ref{result_pitprops} are calculated with the total sparse PC cardinalities being 26, 18 and 15, respectively, and are not included in Figure \ref{piptrop_curve}.

\begin{table}[t]
\centering \caption{Performance comparison of different sparse PCA methods
on pitprops data with different cardinality settings. The best result in each experiment
is highlighted in bold.}
\label{result_pitprops}
\vspace{0.8em}
\footnotesize
\begin{tabular}{ccccccc}
\hline
  & \multicolumn{2}{c}{8-5-6-2-3-2(26)} & \multicolumn{2}{c}{7-4-4-1-1-1(18)} & \multicolumn{2}{c}{7-2-3-1-1-1(15)} \\
 \cline{2-7}
 & RRE & PEV & RRE & PEV & RRE & PEV \\
\hline
  SPCA & 0.4162 & 82.68\% & 0.4448 & 80.22\% & 0.4459 & 80.11\% \\
  DSPCA & 0.4303 & 81.48\% & 0.4563 & 79.18\% & 0.4771 & 77.23\% \\
  PathSPCA & 0.4080 & 83.35\% & 0.4660 & 80.11\% & 0.4457 & 80.13\% \\
  sPCA-rSVD$_{l_0}$ & 0.4139 & 82.87\% & 0.4376 & 80.85\% & 0.4701 & 77.90\% \\
  sPCA-rSVD$_{l_1}$ & 0.4314 & 81.39\% & 0.4427 & 80.40\% & 0.4664 & 78.25\% \\
  sPCA-rSVD$_{\textrm{SCAD}}$ & 0.4306 & 81.45\% & 0.4453 & 80.17\% & 0.4762 & 77.32\% \\
  EMPCA & 0.4070 & 83.44\% & 0.4376 & 80.85\% & 0.4451 & 80.18\% \\
  GPower$_{l_0}$ & 0.4092 & 83.26\% & 0.4400 & 80.64\% & 0.4457 & 80.13\% \\
  GPower$_{l_1}$ & 0.4080 & 83.35\% & 0.4460 & 80.11\% & 0.4457 & 80.13\% \\
  GPower$_{l_{0,m}}$ & 0.4224 & 82.16\% & 0.5089 & 74.10\% & 0.4644 & 78.44\% \\
  GPower$_{l_{1,m}}$ & 0.4187 & 82.46\% & 0.4711 & 77.81\% & 0.4589 & 78.94\% \\
  ALSPCA & 0.4168 & 82.63\% & 0.4396 & 80.67\% & 0.4537 & 79.42\% \\
  ReDaC & \textbf{0.4005} & \textbf{83.50\%} & \textbf{0.4343} & \textbf{81.14\%} & \textbf{0.4420} & \textbf{80.46\%} \\
\hline
\end{tabular}
\end{table}

It can be seen from Table \ref{result_pitprops} that under all cardinality settings of the first $6$ PCs, the proposed ReDaC method always achieves the lowest RRE and highest PEV values among all the competing methods. This means that the ReDaC method is advantageous in both reconstruction-error-minimization and data-variance-maximization viewpoints. Furthermore, from Figure \ref{piptrop_curve}, it is easy to see the superiority of the ReDaC method. In specific, for different number of extracted sparse PC components, the proposed ReDaC method can always get the smallest RRE values and the largest PEV values, as compared with the other utilized sparse PCA methods, in the experiments. This further substantiates the effectiveness of the proposed ReDaC method in both reconstruction-error-minimization and data-variance-maximization views.

\subsubsection{Colon data}

\begin{figure}[t]
\begin{center}
\scalebox{0.49}{\includegraphics[viewport=85 50 963 330]{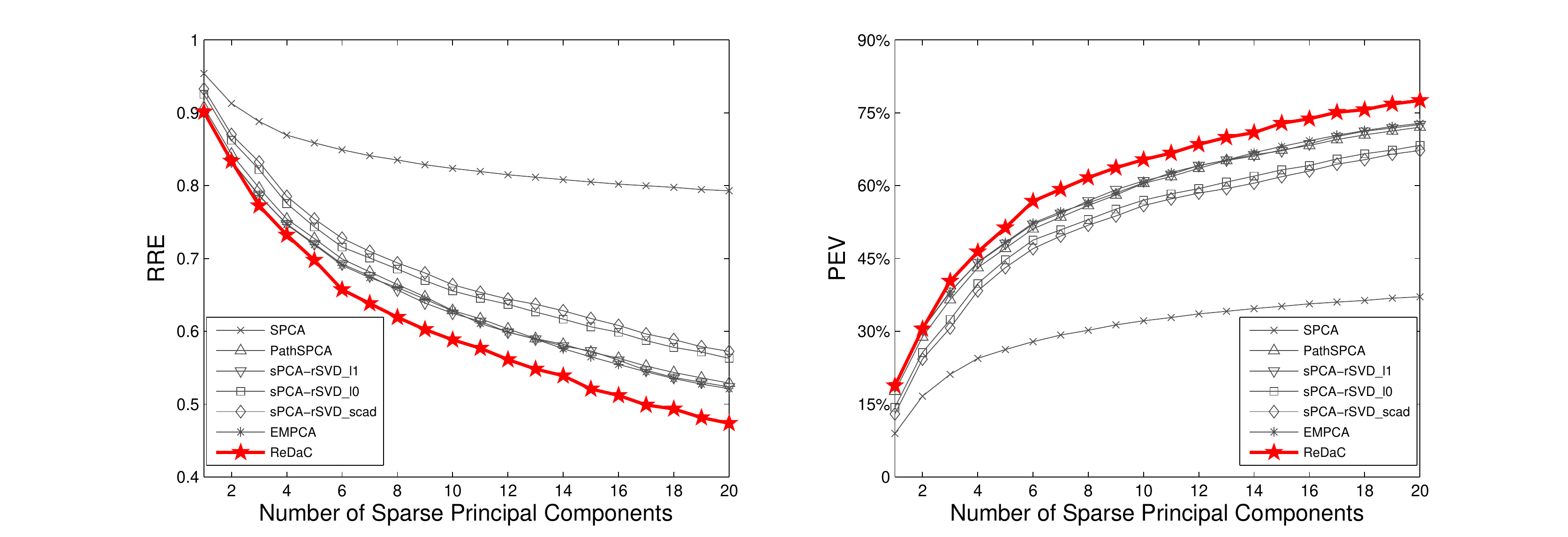}}
\end{center}

\caption{The tendency curves of RRE and PEV with respect to the number of extracted sparse PCs, each with cardinality $50$,
attained by different sparse PCA methods
on colon data.} \label{colon}
\end{figure}

The colon data set \cite{Colon} consists of $62$ tissue samples with
the gene expression profiles of $2000$ genes extracted from DNA micro-array data.
This is a typical data set with high-dimension and low-sample-size property, and is always employed by sparse methods for extracting interpretable information from high-dimensional genes. We thus adopt this data set for evaluation. In specific, $20$ sparse PCs, each with $50$ nonzero loadings, are calculated by different sparse PCA methods, including SPCA \cite{SPCA}, PathSPCA \cite{PathSPCA}, sPCA-rSVD$_{l_0}$, sPCA-rSVD$_{l_1}$, sPCA-rSVD$_{SCAD}$ \cite{rSVD}, EMPCA \cite{EMPCA}, GPower$_{l_{0}}$, GPower$_{l_{1}}$, GPower$_{l_{0,m}}$, GPower$_{l_{1,m}}$ \cite{GPower} and ALSPCA \cite{ALSPCA}, respectively.
Their performance is compared in Table \ref{result_colon} and Figure \ref{colon} in terms of RRE and PEV, respectively.
It should be noted that the DSPCA method has also been tried, while cannot be terminated in a reasonable time in this experiment, and thus we omit its result in the table. Besides, we have carefully tuned the parameters of the GPower methods (including GPower$_{l_{0}}$, GPower$_{l_{1}}$, GPower$_{l_{0,m}}$ and GPower$_{l_{1,m}}$), and can get $20$ sparse PCs with total cardinality around $1000$, similar as the total nonzero elements number of the other utilized sparse PCA methods, while cannot get sparse PC loading sequences each with cardinality $50$ as expected. The results are thus not demonstrated in Figure \ref{colon}.

\begin{table}[t]
\centering \caption{Performance comparison of different sparse PCA methods on colon data. The best results are highlighted in bold.}
\label{result_colon}
\vspace{0.8em}
\footnotesize
\begin{tabular}{ccccc}
\hline
         & SPCA & PathSPCA & sPCA-rSVD$_{l_0}$ & sPCA-rSVD$_{l_1}$  \\
\hline
  RRE. & 0.7892 & 0.5287 & 0.5236 & 0.5628  \\
  PEV. & 37.72\% & 72.05\% & 72.58\% & 68.32\% \\
\hline
\hline
         & sPCA-rSVD$_{\textrm{SCAD}}$ & EMPCA & GPower$_{l_0}$ & GPower$_{l_1}$  \\
\hline
  RRE. & 0.5723 & 0.5211 & 0.5042 & 0.5076  \\
  PEV. & 67.25\% & 72.84\% & 74.56\% & 74.23\% \\
\hline
\hline
         & GPower$_{l_{0,m}}$ & GPower$_{l_{1,m}}$ & ALSPCA & ReDaC \\
\hline
  RRE. & 0.4870 & 0.4904 & 0.5917 & \textbf{0.4737} \\
  PEV. & 76.29\% & 75.95\% & 64.99\% & \textbf{77.56\%} \\
\hline
\end{tabular}
\end{table}

From Table \ref{result_colon}, it is easy to see that the proposed ReDaC method achieves the lowest RRE and highest PEV values, as compared with the other $11$ employed sparse PCA methods. Figure \ref{colon} further demonstrates that as the number of extracted sparse PCs increases, the advantage of the ReDaC method tends to be more dominant than other methods, with respect to both the RRE and PEV criteria. This
further substantiates the effectiveness of the proposed method and implies its potential usefulness in applications with various interpretable components.

\subsection{Nonnegative sparse PCA experiments}

We further testify the performance of the proposed ReDaC method (Algorithm 2) in nonnegative sparse PC extraction. For comparison, two existing methods for nonnegative sparse PCA, NSPCA \cite{NSPCA} and Nonnegative EMPCA (N-EMPCA, briefly) \cite{EMPCA}, are also employed.

\subsubsection{Synthetic toy data}

As the toy data utilized in Section 3.2, we also formulate a Gaussian distribution
$\mathcal{N}(\mathbf{0},\mathbf{\Sigma})$ with mean $\mathbf{0}$ and covariance matrix $\mathbf{\Sigma}=\sum_{j=1}^{10}c_j\mathbf{v}_j\mathbf{v}_j^T\in \mathbb{R}^{10\times10}$.
Both the leading two eigenvectors of $\mathbf{\Sigma}$ are specified as nonnegative and sparse vectors as:
\begin{equation*}
\begin{split}
  &\mathbf{v}_1=(0.474, 0, 0.158, 0, 0.316, 0, 0.791, 0, 0.158, 0)^T,\\
  &\mathbf{v}_2=(0, 0.140, 0, 0.840, 0, 0.280, 0, 0.140, 0, 0.420)^T,
\end{split}
\end{equation*}
and the rest are then generated by applying Gram-Schmidt orthonormalization to $8$ randomly valued
$10$-dimensional vectors. The 10 corresponding eigenvalues $(c_1,c_2,...,c_{10})$ are preset as $(210, 190, 50, 50, 6, 5, 4, 3, 2, 1)$, respectively.
Four series of experiments are designed, each with $1000$ data sets generated from $\mathcal{N}(\mathbf{0},\mathbf{\Sigma})$,
with sample sizes $500$, $1000$, $2000$ and $5000$, respectively. For each experiment, the first two PCs are calculated by the conventional PCA, NSPCA, N-EMPCA and ReDaC methods, respectively. The success times, calculated in the similar way as introduced in Section \ref{sythetic_spca}, of each utilized method on each series of experiments are recorded, as listed in Table \ref{result_simu_identify_nspca}.

\begin{table}[t]
\centering \caption{Performance comparison of success times attained by PCA, NSPCA, N-EMPCA and ReDaC on synthetic toy experiments with different sample sizes. The best results are highlighted in bold.}
\label{result_simu_identify_nspca}
\vspace{0.8em}
\footnotesize
\begin{tabular}{ccccc}
\hline
   & $n=500$ & $n=1000$ & $n=2000$ & $n=5000$ \\
\hline
  PCA & $0$ & $0$ & $0$ & $0$ \\

  NSPCA & $739$ & $948$ & $933$ & $993$ \\

  N-EMPCA & $620$ & $655$ & $631$ & $639$ \\

 ReDaC & $\textbf{835}$ & $\textbf{949}$ & $\textbf{978}$ & $\textbf{1000}$ \\
\hline
\end{tabular}
\end{table}

From Table \ref{result_simu_identify_nspca}, it is seen that the ReDaC method achieves the highest success rates in all experiments. The advantage of the proposed ReDaC method on nonnegative sparse PCA calculation, as compared with the other utilized methods, can thus been verified in these experiments.

\subsubsection{Colon data}

\begin{figure}[t]
\begin{center}
\scalebox{0.61}{\includegraphics[viewport=65 40 700 260]{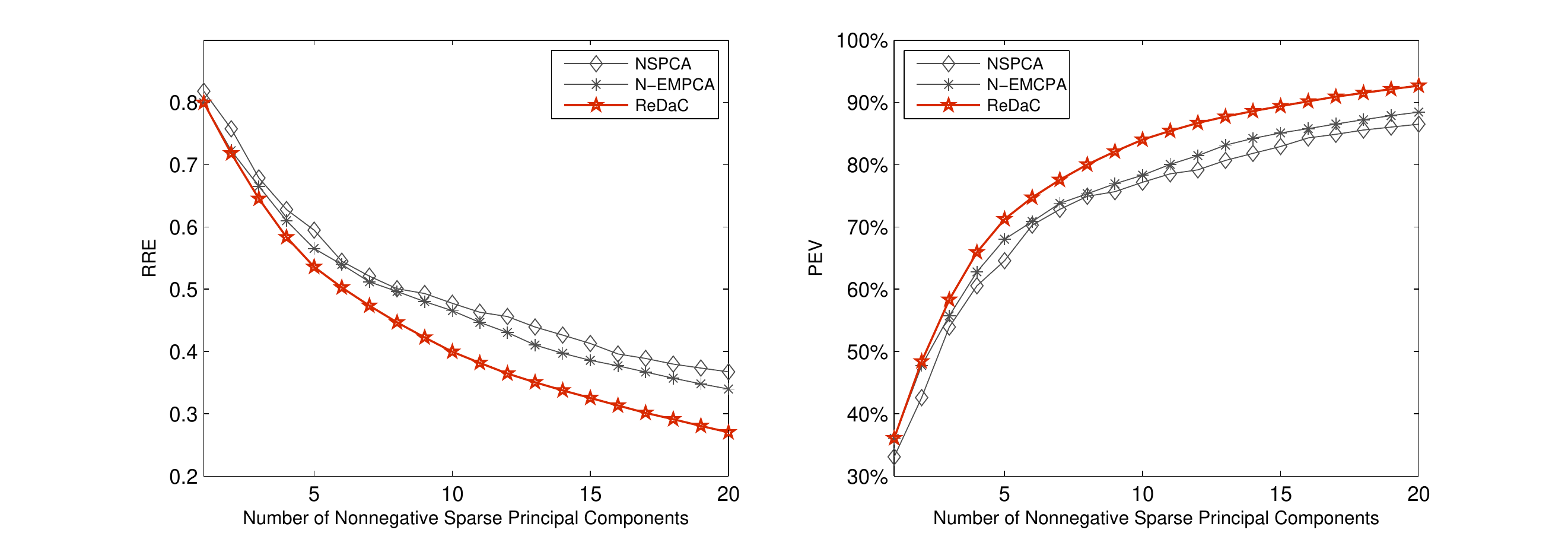}}
\end{center}
\caption{The tendency curves of RRE and PEV, with respect to the number of extracted nonnegative sparse PCs, attained by NSPCA, N-EMPCA and ReDaC on colon data.} \label{colon_curve_nspca}
\end{figure}

The colon data set is utilized again for nonnegative sparse PCA calculation. The NSPCA and N-EMPCA methods are adopted as the competing methods. Since the NSPCA method cannot directly pre-specify the cardinalities of the extracted sparse PCs,
we thus first apply NSPCA on the colon data (with parameters $\alpha=1\times10^6$ and $\beta=1\times10^7$) and then use the cardinalities of the nonnegative sparse PCs attained by this method to preset the N-EMPCA and ReDaC methods for fair comparison. $20$ sparse PCs are computed by the three methods, and the performance is compared in Table \ref{result_colon_nspca} and Figure \ref{colon_curve_nspca}, in terms of RRE and PEV, respectively.

\begin{table}[t]
\centering \caption{Performance comparison of different nonnegative sparse PCA methods on colon data. The best results are highlighted in bold.}
\label{result_colon_nspca}
\vspace{0.8em}
\footnotesize
\begin{tabular}{cccc}
\hline
         & NSPCA & N-EMPCA & ReDaC \\
\hline
  RRE & 0.3674 & 0.3399 & \textbf{0.2706} \\
  PEV & 86.50\% & 88.45\% & \textbf{92.68\%} \\
\hline
\end{tabular}
\end{table}

Just as expected, it is evident that the proposed ReDaC method dominates in both RRE and PEV viewpoints. From Table \ref{result_colon_nspca}, we can observe that our method achieves the lowest RRE and highest PEV on $20$ extracted nonnegative sparse PCs than the other two utilized methods. Furthermore, Figure \ref{colon_curve_nspca} shows that our method is advantageous, as compared with the other methods, for any preset number of extracted sparse PCs, and this advantage tends to be more significant as more sparse PCs are to be calculated. The effectiveness of the proposed method on nonnegative sparse PCA calculation can thus be further verified.

\subsubsection{Application to face recognition}

\begin{figure}[t]
\begin{center}
\scalebox{0.7}{\includegraphics[viewport=5 35 557 220]{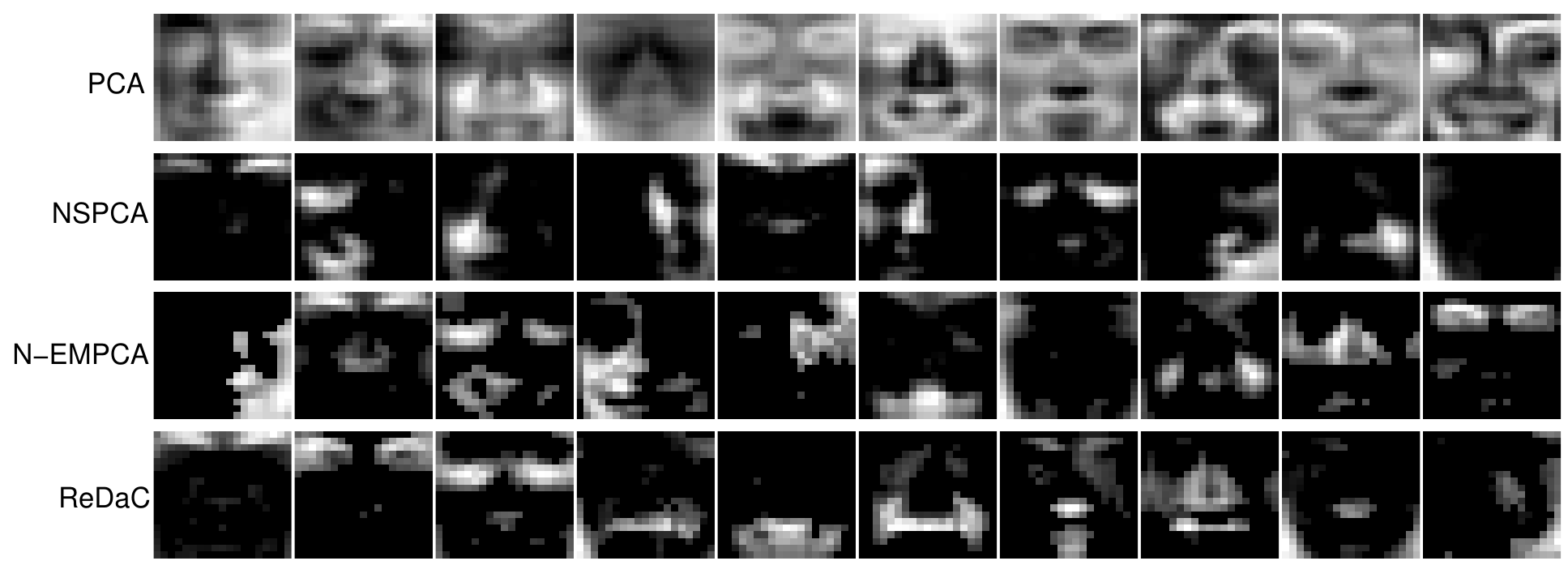}}
\end{center}
\caption{From top row to bottom row: $10$ PCs or nonnegative sparse PCs extracted by PCA, NSPCA, N-EMPCA and ReDaC, respectively.} \label{face_nspca}
\end{figure}

In this section, we introduce the performance of our method in face recognition problem \cite{NSPCA}.
The proposed ReDaC method, together with the conventional PCA, NSPCA and N-EMPCA methods, have been applied to this problem and their performance is compared in this application. The employed data set is the MIT CBCL Face Dataset \#1, downloaded from ``http://cbcl.mit.edu/software-datasets/FaceData2.html''. This data set consists of $2429$ aligned face images and $4548$ non-face images, each with resolution $19\times19$. For each of the four utilized methods, $10$ PC loading vectors are computed on face images, as shown in Figure \ref{face_nspca}, respectively.
For easy comparison, we also list the RRE and PEV values of three nonnegative sparse PCA methods in Table \ref{result_face_nspca}.

\begin{table}[t]
\centering \caption{Performance comparison of different nonnegative sparse PCA methods on MIT CBCL Face Dataset \#1. The best results are highlighted in bold.}
\label{result_face_nspca}
\vspace{0.8em}
\footnotesize
\begin{tabular}{cccc}
\hline
         & NSPCA & N-EMPCA & ReDaC \\
\hline
RRE & 0.6993 & 0.6912 & \textbf{0.6606} \\
PEV & 51.10\% & 52.22\% & \textbf{56.36\%} \\
\hline
\end{tabular}
\end{table}

As depicted in Figure \ref{face_nspca}, the nonnegative sparse PCs obtained by the ReDaC method more clearly exhibit the interpretable features underlying faces, as compared with the other utilized methods, e.g. the first five PCs calculated from our method clearly demonstrate the eyebrows, eyes, cheeks, mouth and chin of faces, respectively. The advantage of the proposed method can further be verified quantitatively by its smallest RRE and largest PEV values, among all employed methods, in the experiment, as shown in Table \ref{result_face_nspca}. The effectiveness of the ReDaC method can thus be substantiated.

To further show the usefulness of the proposed method, we apply it to face classification under this data set as follows. First we randomly choose
$1000$ face images and $1000$ non-face images from MIT CBCL Face Dataset \#1, and take them as the training data and the rest images as testing data. We then extract $10$ PCs by utilizing the PCA, NSPCA, N-EMPCA and ReDaC methods to the training set, respectively. By projecting the training data onto the corresponding $10$ PCs obtained by these four methods, respectively, and then fitting
the linear Logistic Regression (LR) \cite{ESL} model on these dimension-reduced data ($10$-dimensional), we can get a classifier for testing. The classification accuracy of the classifier so obtained on the testing data is then computed, and the results are reported in Table \ref{face_detection_results}. In the table, the classification accuracy attained by directly fitting the LR model on the original training data and testing on the original testing data
is also listed for easy comparison.

\begin{table}[t]
\centering \caption{Performance comparison of the classification accuracy obtained by different nonnegative sparse PCA methods. The best results are highlighted in bold.}
\label{face_detection_results}
\vspace{0.8em}
\footnotesize
\begin{tabular}{cccc}
\hline
     & Face (\%)  & Non-face (\%) & Total (\%) \\
\hline
  LR & 96.71 & 93.57 & 94.47 \\

  PCA + LR & 96.64 & 94.17 & 94.88 \\

  NSPCA + LR & 94.89 & 93.49 & 93.89 \\

  N-EMPCA + LR & 96.71 & 94.39 & 95.06 \\

  ReDaC + LR & \textbf{96.78} & \textbf{94.46} & \textbf{95.84} \\
\hline
\end{tabular}
\end{table}

From Table \ref{face_detection_results}, it is clear that the proposed ReDaC method attains the best performance among all implemented methods, most accurately recognizing both the face images and the non-face images from the testing data. This further implies the potential usefulness of the proposed method in real applications.

\section{Conclusion}

In this paper we have proposed a novel recursive divide-and-conquer method (ReDaC) for sparse PCA problem. The main methodology of the proposed method is to
decompose the original large sparse PCA problem into a series of small sub-problems. We have proved that each of these decomposed sub-problems has a closed-form global solution and can thus be easily solved. By recursively solving these small sub-problems, the original sparse PCA problem can always be
very effectively resolved. We have also shown that the new method converges to a stationary point of the problem, and can be easily extended to other sparse PCA problems with certain constraints, such as nonnegative sparse PCA problem.
The extensive experimental results have validated that our method outperforms current sparse PCA methods in both reconstruction-error-minimization and data-variance-maximization viewpoints.

There are many interesting investigations still worthy to be further explored. For example, when we reformat the square $L_2$-norm error of the sparse PCA model as the $L_1$-norm one, the robustness of the model can always be improved for heavy noise or outlier cases, while the model is correspondingly more difficult to solve. By adopting the similar ReDaC methodology, however, the problem can be decomposed into a series of much simpler sub-problems, which are expected to be much more easily solved than the original model. Besides, although we have proved the convergence of the ReDaC method, we do not know how far the result is from the global optimum of the problem. Stochastic global optimization techniques, such as simulated annealing and evolution computation methods, may be combined with the proposed method to further improve its performance. Also, more real applications of the proposed method are under our current research.

\bibliographystyle{elsarticle-num}
\bibliography{SPCA_ref}

\newpage
\section*{Appendix A. Proof of Theorem 2}

In the following, we denote $\mathbf{w}=\mathbf{E}^{T}\mathbf{u}$, and $%
hard_{\lambda }(\mathbf{w})$ the hard thresholding function, whose $i$-th
element corresponds to $I(|w_{i}|\geq \lambda )w_{i}$, where $w_{i}$
is the $i$-th element of $\mathbf{w}$\ and $I(x)$ (equals $1$ if $x$ is ture%
, and $0$ otherwise)\ is the indicator function

\noindent\textbf{Theorem 2.}
The optimal solution of
\begin{equation*}
\underset{\mathbf{v}}{\max }~~\mathbf{w}^{T}\mathbf{v}~~~~s.t.~~\mathbf{v}^{T}%
\mathbf{v}=1, ~~~\Vert \mathbf{v}\Vert _{0}\leq t,
\end{equation*}%
is given by:%
\begin{equation*}
\mathbf{v}^{\ast }_0(\mathbf{w},t)=\left\{
\begin{array}{cl}
\phi , & t<1\text{,} \\
\frac{hard_{\theta_k}(\mathbf{w})}{\Vert hard_{\theta_k}(\mathbf{w})\Vert _{2}}\mathbf{,}
& k\leq t<k+1\text{ (}k=1,2,\dots ,d-1\text{),} \\
\frac{\mathbf{w}}{\Vert \mathbf{w}\Vert _{2}} & t\geq d\text{.}%
\end{array}%
\right.
\end{equation*}
where $\theta_k$ denotes the $k$-th largest element of $|\mathbf{w}|$.

\begin{pf}
In case of $t<1$, the feasible region of the optimization problem is empty,
and thus the solution of the problem does not exist.

In case of $t\geq d$, the problem is equivalent to
\begin{equation*}
\underset{\mathbf{v}}{\max }~\mathbf{w}^{T}\mathbf{v}~~~~s.t.~~\mathbf{v}^{T}%
\mathbf{v}=1\text{.}
\end{equation*}%
It is then easy to attain the optimum of the problem $\mathbf{v}^{\ast }=\frac{%
\mathbf{w}}{\Vert \mathbf{w}\Vert _{2}}$.

In case of $k\leq t<k+1$ ($k=1,2,\dots ,d-1$), the optimum $\mathbf{v}^{\ast
}$ of the problem is parallel to $\mathbf{w}$ on the $k$-dimensional subspace where the first $k$ largest absolute value of $\mathbf{w}$ are located. Also due to the constraint that $\mathbf{v}^{T}\mathbf{v}=1$, it is then easy to deduce that the optimal solution of the optimization problem is $\frac{hard_{t}(\mathbf{w})}{\Vert hard_{t}(\mathbf{w})\Vert _{2}}$.

The proof is completed.
\end{pf}

\newpage
\section*{Appendix B. Proof of Theorem 3}

We denote $(I_{1},I_{2},\dots ,I_{d})$ the permutation of $(1,2,\dots
,d)$ based on the ascending order of $|\mathbf{w|}=(\mathbf{|}w_{1}\mathbf{|}%
,\mathbf{|}w_{2}\mathbf{|},\dots ,\mathbf{|}w_{d}\mathbf{|})^{T}$, $%
soft_{\lambda }(\mathbf{w})$ the soft thresholding function $sign(\mathbf{%
w})(|\mathbf{w|}-\lambda )_{+}$, $f_{\mathbf{w}}\mathbf{(\lambda )}=\frac{%
soft_{\lambda }(\mathbf{w})}{\left\Vert soft_{\lambda }(\mathbf{w}%
)\right\Vert _{2}}$ and $g_{\mathbf{w}}\mathbf{(\lambda )}=\mathbf{w}^{T}f_{%
\mathbf{w}}\mathbf{(\lambda )}$ throughout the following.

\noindent\textbf{Theorem 3.}
The optimal solution of
\begin{equation*}
\underset{\mathbf{v}}{\max }~~\mathbf{w}^{T}\mathbf{v}~~~~s.t.~~\mathbf{v}^{T}%
\mathbf{v}=1,~~~\Vert\mathbf{v}\Vert _{1}\leq t,
\end{equation*}%
is given by:%
\begin{equation*}
\mathbf{v}^{\ast }_1(\mathbf{w})=\left\{
\begin{array}{cl}
\phi , & t<1, \\
f_{\mathbf{w}}\mathbf{(\lambda }_{k}\mathbf{),} & t\in [\|f_{\mathbf{w}}%
\mathbf{(}|w_{I_{k}}|)\|_1,\|f_{\mathbf{w}}\mathbf{(}|w_{I_{k-1}}|)\|_1)~~(k=2,3,\dots
,d-1), \\
f_{\mathbf{w}}\mathbf{(\lambda }_{1}\mathbf{),} & t\in [\|f_{\mathbf{w}}%
\mathbf{(}|w_{I_{1}}|)\|_1,\sqrt{d}), \\
f_{\mathbf{w}}(0), & t\geq \sqrt{d},%
\end{array}%
\right.
\end{equation*}%
where for $k=1,2,\dots,d-1$,
\begin{equation*}
\mathbf{\lambda }_{k}=\frac{(m-t^{2})(\sum_{i=1}^{m}a_{i})-\sqrt{%
t^{2}(m-t^{2})(m\sum_{i=1}^{m}a_{i}^{2}-(\sum_{i=1}^{m}a_{i})^{2})}}{m(m-t^{2})}\text{,}
\end{equation*}%
where $(a_{1},a_{2},\dots ,a_{m})=(|w_{I_{k}}|,|w_{I_{k+1}}|,\dots
,|w_{I_{d}}|)$, $m=d-k+1$.

\begin{pf}
For any $\mathbf{v}$ located in the feasible region of (\ref{tu0}), it holds that
\begin{equation*}
\sqrt{d}=\sqrt{d\mathbf{v}^{T}\mathbf{v}}\geq \Vert \mathbf{v}\Vert
_{1}\geq \sqrt{\mathbf{v}^{T}\mathbf{v}}=1.
\end{equation*}%
We thus have that if $t<1$, then the optimal solution $\mathbf{v}^{\ast }$ does
not exist since the feasible region of the optimization problem (9) is empty.

If $t\geq \sqrt{d}$, it is easy to see that (\ref{tu0}) is equivalent to
\begin{equation*}
\underset{\mathbf{v}}{\max }~\mathbf{w}^{T}\mathbf{v}~~s.t.\text{ }\mathbf{v}%
^{T}\mathbf{v}=1,
\end{equation*}%
and its optimum is
\begin{equation*}
\mathbf{v}^{\ast }=\frac{\mathbf{w}}{\left\Vert \mathbf{w}\right\Vert _{2}}%
=f_{\mathbf{w}}(0).~
\end{equation*}

We then discuss the case when $t\in \lbrack 1,\sqrt{d})$. Firstly
we deduce the monotonic decreasing property of $h_{\mathbf{w}}(\lambda)=\|f_{\mathbf{w}}\mathbf{(\lambda )}\|_1=\left\|\frac{%
soft_{\lambda }(\mathbf{w})}{\left\Vert soft_{\lambda }(\mathbf{w}%
)\right\Vert _{2}}\right\|_1$ and $g_{\mathbf{w}}\mathbf{(\lambda )}=\mathbf{w}^{T}f_{%
\mathbf{w}}\mathbf{(\lambda )}$ in $\lambda \in
(-\infty ,|w_{I_{d}}|)$ by the following lemmas.

\begin{lemma}
$h_{\mathbf{w}}\mathbf{(\lambda )}$ is monotonically decreasing\ with
respect to $\lambda $ in $(-\infty ,|w_{I_{d}}|)$.
\end{lemma}

\begin{pf}
First, we prove that $h_{\mathbf{w}}\mathbf{(\lambda )}$ is monototically
decreasing with $\lambda $ $\in \lbrack \mathbf{|}w_{I_{k-1}}\mathbf{|},%
\mathbf{|}w_{I_{k}}\mathbf{|}),$ $k=2,3,\dots ,d$ and $(-\infty ,|w_{I_{1}}|)$.

It is easy to see that for $\lambda $ $\in \lbrack \mathbf{|}w_{I_{k-1}}%
\mathbf{|},\mathbf{|}w_{I_{k}}\mathbf{|}),$ $k=2,3,\dots ,d$ and $(-\infty
,|w_{I_{1}}|)$,%
\begin{equation*}
h_{\mathbf{w}}\mathbf{(\lambda )}=\frac{\sum_{i=k}^{d}(\mathbf{|}w_{I_{i}}%
\mathbf{|}-\lambda )}{\sqrt{\sum_{i=k}^{d}(\mathbf{|}w_{I_{i}}\mathbf{|}%
-\lambda )^{2}}}.
\end{equation*}%
Then we have%
\begin{eqnarray*}
h_{\mathbf{w}}^{\prime }\mathbf{(\lambda )} &=&\frac{-(d-k+1)\sqrt{%
\sum_{i=k}^{d}(\mathbf{|}w_{I_{i}}\mathbf{|}-\lambda )^{2}}+\frac{%
\sum_{i=k}^{d}(\mathbf{|}w_{I_{i}}\mathbf{|}-\lambda )}{\sqrt{\sum_{i=k}^{d}(%
\mathbf{|}w_{I_{i}}\mathbf{|}-\lambda )^{2}}}\sum_{i=k}^{d}(\mathbf{|}%
w_{I_{i}}\mathbf{|}-\lambda )}{\sum_{i=k}^{d}(|w_{I_{i}}|-\lambda )^{2}}
\\
&=&\left( \sum_{i=k}^{d}(\mathbf{|}w_{I_{i}}\mathbf{|}-\lambda )^{2}\right)
^{-3/2}\left( -(d-k+1)\sum_{i=k}^{d}(\mathbf{|}w_{I_{i}}\mathbf{|}-\lambda
)^{2}+\left( \sum_{i=k}^{d}(\mathbf{|}w_{I_{i}}\mathbf{|}-\lambda )\right)
^{2}\right) .
\end{eqnarray*}%
It is known that for any number sequence $s_{1},s_{2},\dots ,s_{n}$, it
holds that
\begin{equation*}
\left( \sum_{i=1}^{n}s_{i}\right) ^{2}\leq n\sum_{i=1}^{n}s_{i}^{2}.
\end{equation*}%
Thus we have
\begin{equation*}
h_{\mathbf{w}}^{\prime }\mathbf{(\lambda )\leq 0}
\end{equation*}%
for $\lambda $ $\in \lbrack \mathbf{|}w_{I_{k-1}}\mathbf{|},\mathbf{|}%
w_{I_{k}}\mathbf{|}),$ $k=2,3,\dots ,d$ and $(-\infty ,|w_{I_{1}}|)$.
Since $h_{\mathbf{w}}\mathbf{(\lambda )}$ is obviously a continuous function
in $(-\infty ,\mathbf{|}w_{I_{d}}\mathbf{|})$, it can be easily deduced that
$h_{\mathbf{w}}\mathbf{(\lambda )}$ is monotonically decreasing\ in the
entire set $(-\infty ,\mathbf{|}w_{I_{d}}\mathbf{|})$ with respect to $\lambda$.

The Proof is completed.
\end{pf}

Based on Lemma 1, It is easy to deduce that the range of $h_{\mathbf{w}}%
\mathbf{(\lambda )}$ for $\lambda \in (-\infty ,\mathbf{|}w_{I_{d}}\mathbf{|}%
)$ is $[1,\sqrt{d})$, since $\underset{\mathbf{\lambda \rightarrow -\infty }}%
{\lim }h_{\mathbf{w}}(\mathbf{\lambda })=\sqrt{d}$ and $h_{\mathbf{w}}(%
\mathbf{\lambda })=1$ for $\mathbf{\lambda \in \lbrack |}%
w_{I_{d-1}}|,|w_{I_{d}}\mathbf{|})\mathbf{.}$

The following lemma shows the monotonic decreasing property of $g_{\mathbf{w}}\mathbf{(\lambda )}$.

\begin{lemma}
$g_{\mathbf{w}}\mathbf{(\lambda )}$ is monotonically decreasing\ with
respect to $\lambda \in (-\infty ,|w_{I_{d}}|)$.
\end{lemma}

\begin{pf}
Please see \cite{RSPCA} for the proof.
\end{pf}

The next lemma proves that the optimal solution $\mathbf{v}^{\ast }$ can be
expressed as $f_{\mathbf{w}}\mathbf{(\lambda ^{\ast })}$.

\begin{lemma}
The optimal solution of (\ref{tu0}) is of the expression $\mathbf{v}^{\ast }=f_{%
\mathbf{w}}\mathbf{(\lambda ^{\ast })}$ for $t\in \lbrack 1,\sqrt{d})$ on
some $\lambda ^{\ast }\in (-\infty ,|w_{I_{d}}|)$.
\end{lemma}

\begin{pf}
Please see \cite{RSPCA, PMD} for the proof.
\end{pf}

Lemmas 1-3 imply that the optimal solution of (\ref{tu0}) is attained at $\lambda
^{\ast }$ where $\Vert f_{\mathbf{w}}\mathbf{(\lambda ^{\ast })}\Vert _{1}=t$
holds. The next lamma presents the closed-form solution of this equation.

\begin{lemma}
The solutuion of $\Vert f_{\mathbf{w}}\mathbf{(\lambda )}\Vert _{1}=t$ for $%
t\in [\|f_{\mathbf{w}}\mathbf{(}|w_{I_{k}}|)\|_1,\|f_{\mathbf{w}}\mathbf{(}%
|w_{I_{k-1}}|)\|_1),$ $(k=2,3,\dots ,d-1)$, or $t\in [\|f_{\mathbf{w}}\mathbf{(}%
|w_{I_{1}}|)\|_1,\sqrt{d})$ is
\begin{equation*}
\mathbf{\lambda }_{k}=\frac{(m-t^{2})(\sum_{i=1}^{m}a_{i})-\sqrt{%
t^{2}(m-t^{2})(m\sum_{i=1}^{m}a_{i}^{2}-(\sum_{i=1}^{m}a_{i})^{2})}}{m(m-t^{2})}\text{,}
\end{equation*}%
where $(a_{1},a_{2},\dots ,a_{m})=(|w_{I_{k}}|,|w_{I_{k+1}}|,\dots
,|w_{I_{d}}|)$ and $m=d-k+1.$
\end{lemma}

\begin{pf}
Let's transform the equation%
\begin{equation}\label{eqn_lambda1}
\Vert f_{\mathbf{w}}\mathbf{(\lambda )}\Vert _{1}=\frac{\sum_{i=k}^{d}(%
\mathbf{|}w_{I_{i}}\mathbf{|}-\lambda )}{\sqrt{\sum_{i=k}^{d}(\mathbf{|}%
w_{I_{i}}\mathbf{|}-\lambda )^{2}}}=\frac{\sum_{i=1}^{m}(a_{i}-\lambda )}{%
\sqrt{\sum_{i=1}^{m}(a_{i}-\lambda )^{2}}}=t
\end{equation}%
as the following expression%
\begin{equation*}
(\sum_{i=1}^{m}a_{i}-m\lambda )^{2}=t^{2}\sum_{i=1}^{m}(a_{i}-\lambda )^{2}.
\end{equation*}%
Then we can get the quadratic equation with respect to $\lambda $ as:
\begin{equation}\label{eqn_lambda2}
m(m-t^{2})\lambda ^{2}-2(m-t^{2})(\sum_{i=1}^{m}a_{i})\lambda
+(\sum_{i=1}^{m}a_{i})^{2}-t^{2}\sum_{i=1}^{m}a_{i}^{2}=0.
\end{equation}%
We first claim that $t^2< m$ for $t\in [\|f_{\mathbf{w}}\mathbf{(}|w_{I_{k}}|)\|_1,\|f_{\mathbf{w}}\mathbf{(}%
|w_{I_{k-1}}|)\|_1)$, $k=2,3,\dots ,d-1$, or $t\in [\|f_{\mathbf{w}}\mathbf{(}%
|w_{I_{1}}|)\|_1,\sqrt{d})$. In fact, by the definition of $f_{\mathbf{w}}\mathbf{(\lambda )}$, we have that
\begin{equation*}
  \begin{split}
    t&<\|f_{\mathbf{w}}\mathbf{(}|w_{I_{k-1}}|)\|_1=\frac{\sum_{i=1}^{m}(a_{i}-|w_{I_{k-1}}| )}{\sqrt{\sum_{i=1}^{m}(a_{i}-|w_{I_{k-1}}| )^{2}}}\\
    &\leq \left(m\sum_{i=1}^{m}\left(\frac{(a_{i}-|w_{I_{k-1}}| )}{\sqrt{\sum_{i=1}^{m}(a_{i}-|w_{I_{k-1}}| )^{2}}}\right)^2\right)^{\frac{1}{2}}\\
    &=\sqrt{m},
  \end{split}
\end{equation*}
for $t\in [\|f_{\mathbf{w}}\mathbf{(}|w_{I_{k}}|)\|_1,\|f_{\mathbf{w}}\mathbf{(}|w_{I_{k-1}}|)\|_1)$, $k=2,\dots ,d-1$, and
\begin{equation*}
  \begin{split}
    t&<\|f_{\mathbf{w}}\mathbf{(}|\sqrt{d}|)\|_1=\frac{\sum_{i=1}^{d}(a_{i}-|\sqrt{d}| )}{\sqrt{\sum_{i=1}^{d}(a_{i}-|\sqrt{d}| )^{2}}}\\
    &\leq \left(d\sum_{i=1}^{d}\left(\frac{(a_{i}-|\sqrt{d}| )}{\sqrt{\sum_{i=1}^{d}(a_{i}-|\sqrt{d}| )^{2}}}\right)^2\right)^{\frac{1}{2}}\\
    &=\sqrt{d}=\sqrt{m},
  \end{split}
\end{equation*}
for $t\in [\|f_{\mathbf{w}}\mathbf{(}|w_{I_{1}}|)\|_1,\sqrt{d})$. Then it can be seen that the discriminant of equation (\ref{eqn_lambda2})
\begin{equation*}
\Delta=t^{2}(m-t^{2})(m\sum_{i=1}^{m}a_{i}^{2}-(\sum_{i=1}^{m}a_{i})^{2})\geq0,
\end{equation*}
using the fact that $(\sum_{i=1}^{m}a_{i})^{2}\leq m\sum_{i=1}^{m}a_{i}^{2}$. Therefore, the solutions of equation (\ref{eqn_lambda2}) can be expressed as
\begin{equation*}
\lambda =\frac{(m-t^{2})(\sum_{i=1}^{m}a_{i})\pm \sqrt{t^{2}(m-t^{2})(m\sum_{i=1}^{m}a_{i}^{2}-(\sum_{i=1}^{m}a_{i})^{2})}}{m(m-t^{2})}.
\end{equation*}%
It holds that
\begin{eqnarray*}
\lambda ^{+} &=&\frac{(m-t^{2})(\sum_{i=1}^{m}a_{i})+ \sqrt{t^{2}(m-t^{2})(m\sum_{i=1}^{m}a_{i}^{2}-(\sum_{i=1}^{m}a_{i})^{2})}}{m(m-t^{2})}\\
&\geq &\frac{(m-t^{2})(\sum_{i=1}^{m}a_{i})}{m(m-t^{2})} \\
&=&\frac{\sum_{i=1}^{m}a_{i}}{m}(=\frac{\sum_{i=k}^{d}|w_{I_{i}}|}{d-k+1}) \\
&\geq &|w_{I_{k}}|.
\end{eqnarray*}%
If $\lambda ^{+}>|w_{I_{k}}|$, since $\lambda\leq|w_{I_{k}}|$ required by
equation (\ref{eqn_lambda1}), then
\begin{equation*}
\lambda _{k}=\lambda ^{-}=\frac{(m-t^{2})(\sum_{i=1}^{m}a_{i})-\sqrt{t^{2}(m-t^{2})(m\sum_{i=1}^{m}a_{i}^{2}-(\sum_{i=1}^{m}a_{i})^{2})}}{m(m-t^{2})}.
\end{equation*}%
Otherwise, if $\lambda ^{+}=|w_{I_{k}}|$, then it holds that $(\sum_{i=1}^{m}a_{i})^{2}= m\sum_{i=1}^{m}a_{i}^{2}$, which naturally leads to
$\lambda_k=\lambda ^{+}=\lambda ^{-}$.

The proof is then completed.
\end{pf}

Based on the above Lemmas 1-4, the conclusion of Theorem 3 can then be obtained.
\end{pf}

\newpage
\section*{Appendix C. Proof of Theorem 5}

\noindent\textbf{Theorem 5.}
The global optimal solution to (\ref{nsv}) is $\mathbf{v}%
_{p}^{\ast }((\mathbf{w})_{+},t)$ ($p=0,1$), where $\mathbf{w=E}^{T}\mathbf{u}$,
and$\ \mathbf{v}_{0}^{\ast}(\cdot,\cdot )$ and $\mathbf{v}_{1}^{\ast }(\cdot,\cdot )$\
are defined in Theorem 2 and Theorem 3, respectively.

It is easy to prove this theorem based on the following lemma.

\begin{lemma}
Assume that there is at least one element of $\mathbf{w}$ is positive, then the
optimization problem
\begin{equation*}
(P1)~~~\underset{\mathbf{v}}{\max }\text{ }\mathbf{w}^{T}\mathbf{v}~~~s.t.~\
\mathbf{v}^{T}\mathbf{v}=1,~\Vert \mathbf{v}\Vert _{p}\leq t,~\mathbf{v}\succeq
0,
\end{equation*}%
can be equivalently soved by
\begin{equation*}
(P2)~~~\underset{\mathbf{v}}{\max }\text{ }(\mathbf{w})_{+}^{T}\mathbf{v}%
~~~s.t.~\ \mathbf{v}^{T}\mathbf{v}=1,~\Vert \mathbf{v}\Vert _{p}\leq t,
\end{equation*}%
where $p$ is $0$ or $1$.
\end{lemma}

\begin{pf}
Denote the optimal solutions of ($P1$) and ($P2$) as $\mathbf{v1}$ and $%
\mathbf{v2}$, respectively.

First, we prove that $\mathbf{w}^{T}\mathbf{v1}\geq \mathbf{w}^{T}\mathbf{v2}$. Based on
Theorem 2 and 3, the elements of $\mathbf{v2}$ are of the same signs (or
zeros) with the corresponding ones of $(\mathbf{w})_{+}$. This means that $%
\mathbf{v2}\succeq 0$ natrually holds. That is, $\mathbf{v2}$ belongs to the
feasible region of $(P1)$. Since $\mathbf{v1}$ is the optimum of $(P1)$, we
have $\mathbf{w}^{T}\mathbf{v1\geq \mathbf{w}}^{T}\mathbf{v2}$.

Then we prove that $\mathbf{w}^{T}\mathbf{v1}\leq \mathbf{w}^{T}\mathbf{v2}$ through the
following three steps.

(C1): The nonzero elements of $\mathbf{v1=}%
(v_{1}^{(1)},v_{2}^{(1)},...,v_{d}^{(1)})$ lie on the positions where the
nonnegative entries of $\mathbf{w}$ are located.

If all elements of $\mathbf{w}$ are nonnegative, then (C1) is
evidently satisfied.

Otherwise, there is an element, denoted as the $i$-th element $w_{i}$ of $%
\mathbf{w}$, is negative and the corresponding element, $v_{i}^{(1)}$, of $%
\mathbf{v1}$ is nonzero (i.e. positive). We further pick up a nonnegative
element, denoted as $w_{j}$, from $\mathbf{w}$. Then we can construct a new $%
d$-dimensional vector $\widetilde{\mathbf{v}}=(\tilde{v}_{1},\tilde{v}%
_{2},...,\tilde{v}_{i})$ as%
\begin{equation*}
\tilde{v}_{k}=%
\begin{cases}
0, & k=i, \\
\sqrt{(v_{i}^{(1)})^{2}+(v_{j}^{(1)})^{2}}, & k=j, \\
v_{k}^{(1)}, & k\neq i,j.%
\end{cases}%
\end{equation*}%
Then we have

\begin{equation*}
\begin{split}
\mathbf{w}^{T}\tilde{\mathbf{v}}& =\sum_{k}w_{k}v_{k}=w_{j}\sqrt{%
(v_{i}^{(1)})^{2}+(v_{j}^{(1)})^{2}}+\sum_{k\neq i,j}w_{k}v_{k}^{(1)} \\
& >w_{i}v_{i}^{(1)}+w_{j}v_{j}^{(1)}+\sum_{k\neq
i,j}w_{k}v_{k}^{(1)} \\
& =\mathbf{w}^{T}\mathbf{v1}.
\end{split}%
\end{equation*}%
We get the inequality by the fact that $w_{j}\sqrt{%
(v_{i}^{(1)})^{2}+(v_{j}^{(1)})^{2}}\geq w_{j}v_{j}^{(1)}$ and $%
0>w_{i}v_{i}^{(1)}$. This is contradict to the fact that $\mathbf{v1}$ is
the optimal solution of ($P1$), noting that $\|\tilde{\mathbf{v}}\|_p\leq\|\mathbf{v}\|_p\leq t$.

The conclusion (C1) is then proved.

(C2): The nonzero elements of $\mathbf{v2=}%
(v_{1}^{(2)},v_{2}^{(2)},...,v_{d}^{(2)})$ lie on the positions where the
nonzero entries of $(\mathbf{w})_{+}$ are located.

Denote $(\mathbf{w})_{+}=(w_{1}^{+},w_{2}^{+},...,w_{d}^{+})$. If all
elements of $(\mathbf{w})_{+}$ are positive, then (C2) is
evidently satisfied.

Otherwise, let $w_{i}^{+}$ be a zero element of $\mathbf{w}$ and the
corresponding element, $v_{i}^{(2)}$, of $\mathbf{v2}$ is nonzero, and let $%
w_{j}^{+}$ be a positive element of $\mathbf{w}$. Then we can construct a
new $d$-dimensional vector $\overline{\mathbf{v}}=(\overline{v}_{1},%
\overline{v}_{2},...,\overline{v}_{i})$ as%
\begin{equation*}
\overline{v}_{k}=%
\begin{cases}
0, & k=i, \\
\sqrt{(v_{i}^{(2)})^{2}+(v_{j}^{(2)})^{2}}, & k=j, \\
v_{k}^{(2)}, & k\neq i,j.%
\end{cases}%
\end{equation*}%
Then we have

\begin{equation*}
\begin{split}
(\mathbf{w})_{+}^{T}\overline{\mathbf{v}}& =\sum_{k}w_{k}^{+}\overline{v}%
_{k}=w_{j}^{+}\sqrt{(v_{i}^{(2)})^{2}+(v_{j}^{(2)})^{2}}+\sum_{k\neq
i,j}w_{k}^{+}v_{k}^{(2)} \\
& >w_{i}^{+}v_{i}^{(2)}+w_{j}^{+}v_{j}^{(2)}+\sum_{k\neq
i,j}w_{k}^{+}v_{k}^{(1)} \\
& =(\mathbf{w})_{+}^{T}\mathbf{v}2.
\end{split}%
\end{equation*}%
We get the first inequality by the fact that $w_{j}^{+}\sqrt{%
(v_{i}^{(2)})^{2}+(v_{j}^{(2)})^{2}}>w_{j}^{+}v_{j}^{(2)}$ and $%
0=w_{i}^{+}v_{i}^{(2)}$. This is contradict to the fact that $\mathbf{v2}$
is the optimal solution of ($P2$), noting that $\|\tilde{\mathbf{v}}\|_p\leq\|\mathbf{v}\|_p\leq t$.

The conclusion (C2) is then proved.

(C3): We can then prove that $\mathbf{w}^{T}\mathbf{v1}\leq \mathbf{w}^{T}\mathbf{v2}$
based on the conclusions (C1) and (C2) as follows:
\begin{equation*}
\mathbf{w}^{T}\mathbf{v1}= \mathbf{(w)}_{+}^{T}\mathbf{v1}\leq \mathbf{(w)}_{+}^{T}\mathbf{v2=w}^{T}%
\mathbf{v2}.
\end{equation*}%
In the above equation, the first equality is conducted by (C1), the
second inequality is based on the fact that $\mathbf{v2}$ is the optimal
solution of ($P2$), and the third equality is followed by (C2).

Thus it holds that $\mathbf{w}^{T}\mathbf{v1}=\mathbf{w}^{T}\mathbf{v2}$. This implies that the
optimization problem ($P1$) can be equivalently solved by ($P2$).
\end{pf}

\end{document}